\documentclass[11pt]{article}
\usepackage[top=1in,bottom=1in,left=1in,right=1in]{geometry}

\usepackage{amsthm}
\usepackage{graphicx}
\usepackage{epstopdf}
\usepackage{algorithmic}
\usepackage{amsmath}
\usepackage{amssymb}
\usepackage{adjustbox}
\usepackage{hyperref}
\usepackage[T1]{fontenc}
\usepackage{soul,xcolor}
\usepackage{cases}
\usepackage{booktabs}
\usepackage{float}
\usepackage{bm}
\usepackage{setspace}
\usepackage{makecell}
\usepackage{mathtools}
\usepackage{enumerate,enumitem}
\usepackage{lineno}
\usepackage{multirow}

\DeclareMathOperator{\diag}{diag}

\title{MOLA: Enhancing Industrial Process Monitoring Using Multi-Block Orthogonal Long Short-Term Memory Autoencoder}

\author{Fangyuan Ma$^{1,2}$, Cheng Ji$^{2}$, Jingde Wang$^{2}$, Wei Sun$^{2}$, Xun Tang$^{3,\dagger}$, Zheyu Jiang$^{1,\star}$}

\date{
    \normalsize $^1$School of Chemical Engineering, Oklahoma State University, 420 Engineering North, Stillwater, Oklahoma, 74078 USA\\
    $^2$College of Chemical Engineering, Beijing University of Chemical Technology, Beijing, 100029 China\\
    $^3$Cain Department of Chemical Engineering, Louisiana State University, Baton Rouge, Louisiana, 70803 USA
}

\begin{document}
\maketitle
\vspace{-2em}
\noindent Correspondence: $^\star$\texttt{zheyu.jiang@okstate.edu} (Z.J.), $^\dagger$\texttt{xuntang@lsu.edu} (X.T.)

\begin{abstract}
In this work, we introduce MOLA: a Multi-block Orthogonal Long short-term memory Autoencoder paradigm, to conduct accurate, reliable fault detection of industrial processes. To achieve this, MOLA effectively extracts dynamic orthogonal features by introducing an orthogonality-based loss function to constrain the latent space output. This helps eliminate the redundancy in the features identified, thereby improving the overall monitoring performance. On top of this, a multi-block monitoring structure is proposed, which categorizes the process variables into multiple blocks by leveraging expert process knowledge about their associations with the overall process. Each block is associated with its specific Orthogonal Long short-term memory Autoencoder model, whose extracted dynamic orthogonal features are monitored by distance-based Hotelling's $T^2$ statistics and quantile-based cumulative sum (CUSUM) designed for multivariate data streams that are nonparametric and heterogeneous. Compared to having a single model accounting for all process variables, such a multi-block structure significantly improves overall process monitoring performance, especially for large-scale industrial processes. Finally, we propose an adaptive weight-based Bayesian fusion (W-BF) framework to aggregate all block-wise monitoring statistics into a global statistic that we monitor for faults. Fault detection speed and accuracy are improved by assigning and adjusting weights to blocks based on the sequential order in which alarms are raised. We demonstrate the efficiency and effectiveness of our MOLA framework by applying it to the Tennessee Eastman Process and comparing the performance with various benchmark methods.

\end{abstract}


\section{Introduction}

Effective, reliable process monitoring is essential to ensuring process safety, improving product quality, and reducing operating costs of industrial systems as they continue to expand in scale and complexity \cite{amin2018process}. Nowadays, modern chemical plants are equipped with numerous sensors connected to Distributed Control Systems (DCSs), continuously generating massive process data that can be leveraged for data-driven process monitoring in real time \cite{nawaz2022review}. Traditional methods for process monitoring encompass Principal Component Analysis (PCA), Partial Least Squares (PLS), and Independent Component Analysis (ICA), among many others \cite{qin2012survey}. The idea behind these fault detection methods largely falls in extracting underlying features that characterize process states (e.g., faulty vs. non-faulty) from historical process data and monitoring changes in these extracted features \cite{li2022nonlinear}. For instance, PCA leverages linear orthogonal transformations to extract key process features, projecting them in a principal component subspace and a residual subspace, followed by developing a monitoring statistic for each subspace for fault detection \cite{dong2018novel}. Nevertheless, the relationships among process variables being monitored in modern industrial systems are often highly nonlinear, and conventional linear methods such as PCA face challenges in effectively capturing these nonlinear relationships. To overcome the intrinsic limitations of conventional linear methods, kernel-based methods, such as kernel PCA, have been proposed \cite{bounoua2021fault, pilario2019review}. While these kernel-based methods can extract nonlinear relationships, identifying and computing the kernel functions can be time-consuming, limiting their capabilities in real-world applications that demand fast real-time fault detection \cite{li2022nonlinear}. Furthermore, compared to standard PCA, kernel-based methods exhibit greater sensitivity to noise and outliers \cite{tan2020monitoring}, potentially deteriorating monitoring accuracy.

Leveraging the recent breakthroughs in deep learning, Artificial Neural Networks (ANNs), which consist of multiple fully connected layers and nonlinear activation functions, have achieved remarkable successes in extracting complex nonlinear features among process variables \cite{abiodun2018state} for process monitoring. Among prevailing ANN-based methods, Multi-layer Perceptron (MLP), Convolutional Neural Network (CNN), and Recurrent Neural Network (RNN) are some of the most effective and widely-used deep learning architectures for process monitoring \cite{wu2018deep, arunthavanathan2021deep,heo2018fault}. Since traditional ANN-based process monitoring methods are essentially supervised classification methods \cite{yang2022autoencoder}, their monitoring performance relies on the availability of a large amount of labeled data, especially faulty data which are typically limited and hard to acquire in practice \cite{ji2022review}.

\begin{figure}[ht!]
    \centering
    \includegraphics[width=0.7\textwidth]{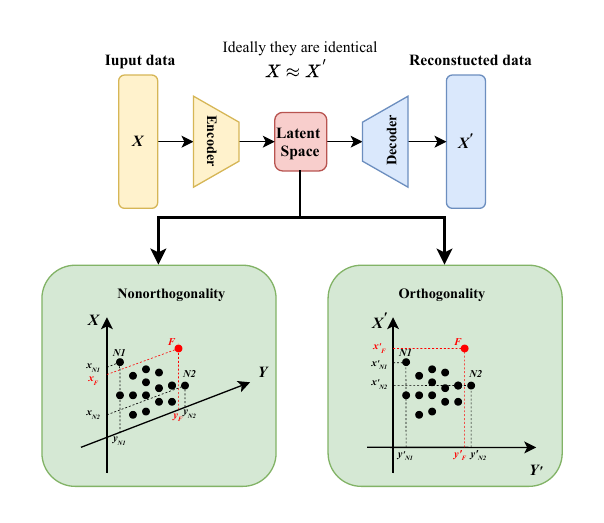}
    \vspace{-2em}
    \caption{Illustration of autoencoder structure and feature extraction.}
    \label{fig_AE}
\end{figure}

Meanwhile, unsupervised methods, such as the autoencoder (AE), have gathered increasing attention due to their ability to extract features from unlabeled data, thereby presenting a more viable alternative for process monitoring \cite{ji2022review}. As illustrated in Figure \ref{fig_AE}, the core structure of an AE comprises an input layer, an encoder layer, a latent space, a decoder layer, and an output layer. Specifically, an encoder maps the original input data into its latent space consisting of codes (or latent variables) that effectively capture and retain the key data representations. A decoder is then employed to accurately reconstruct the original information from these lower-dimensional embeddings, aiming to reproduce data that are indistinguishable from the original input data \cite{fan2017autoencoder}.

In traditional AE-based process monitoring methods, the reconstruction error is used as the primary objective during training and is monitored for fault detection during deployment \cite{qian2022review}. However, this approach does not explicitly make use of the lower-dimensional embeddings which inherently represent the underlying process dynamics. Furthermore, solely minimizing the reconstruction error may introduce redundancy among extracted features (see Figure \ref{fig_AE}). This is illustrated in Figure \ref{fig_AE}, where the points shown represent the projections of the original input data onto a two-dimensional latent space after passing through the encoder layer. Here, the red point $F$ denotes the projection of a faulty data sample, whereas the rest represent non-faulty data samples. When the extracted features contain redundancy, the two dimensions of the latent space are not orthogonal to each other. In this case, the projections of $F$ onto $X$- and $Y$-axes will fall in the range of non-faulty data sample projections, thereby making fault detection more challenging. On the other hand, when the extracted features contain no redundancy, the latent variables are orthogonal to one another. In the same example, the projection of $F$ onto the $X'$-axis now falls outside of the range of non-faulty data sample projections.

Based on this observation, Orthogonal Autoencoder (OAE), which introduces a term characterizing orthogonality of latent space outputs in the loss function, was proposed to extract features that are inherently independent or nonredundant \cite{wang2019clustering}. Cacciarelli et al. then applied the OAE to process monitoring applications by monitoring Hotelling's $T^2$ statistics of orthogonal latent features \cite{cacciarelli2022novel}. While OAE effectively improves the fault detection performance compared to traditional AE, it still has several limitations. First, actual industrial systems typically involve dynamic behaviors \cite{md2020review}, which cannot be captured by existing OAE-based framework. Second, Hotelling's $T^2$ statistic may not be suited for capturing changes and shifts in the distribution of latent features \cite{ji2024orthogonal}. It also does not explicitly account for the cumulative effects of process anomalies, which can play a crucial role in detecting certain types of faults promptly \cite{nawaz2021improved}.

Furthermore, one of the unique characteristics of modern chemical process systems is that they typically comprise multiple heavily integrated (via mass, energy, and information flows) yet relatively autonomous subsystems, each with specific process functions such as raw material processing, reaction, and product separations. These subsystems are further disaggregated into smaller unit operations, such as reactors, heat exchangers, and distillation columns. Such structural complexity and hierarchy can pose significant challenges to conventional process monitoring paradigms that rely on a single model to monitor the entire process \cite{ge2013distributed}. As the number of process variables being monitored increases, the number of hyperparameters in the deep learning model increases exponentially, significantly amplifying the training complexity. To address this challenge, a multi-block process monitoring methodology has been proposed for large-scale industrial process systems \cite{huang2019fault}. The idea is to categorize process variables into various blocks based on the variables' associations with the overall process and their relevance with other process variables, followed by building a process monitoring model for each block. The entire process will be monitored by integrating these block-wise process monitoring models via a data fusion mechanism \cite{zhai2020multi}. Conventional data fusion techniques, such as Bayesian fusion, are static in nature and treat the monitoring statistics from different blocks equally \cite{li2020plant}. However, in reality, process anomalies often stem from one block and propagate/spread to others as time progresses, making conventional data fusion techniques inadequate and less effective.

To address the aforementioned challenges, we propose a novel process monitoring framework based on Multi-block Orthogonal Long short-term memory Autoencoder (MOLA). As a significant variant of traditional autoencoders, Long Short-Term Memory Autoencoder (LSTM AE) implements the LSTM architecture in both the encoder and decoder layers. This allows the extraction of dynamic process features from the time series data. On top of this, we propose the Orthogonal LSTM Autoencoder (OLAE), which incorporates an orthogonality-based loss function to constrain the latent space output. We also adopt the multi-block monitoring methodology to assign all process variables into several blocks based on process knowledge. A local OLAE model is developed for each block. To effectively detect anomalies of the orthogonal latent features in each block, in addition to Hotelling's $T^2$ approach, we also incorporate a quantile-based multivariate cumulative sum (CUSUM) process monitoring method \cite{ye2022generic}, a state-of-the-art approach to monitor high-dimensional data streams that are nonparametric (i.e., data streams do not necessarily follow any specific distribution) and heterogeneous (i.e., data streams do not necessarily follow the same distribution) \cite{jiang2023online}. The use of quantile-based multivariate CUSUM successfully overcomes the barrier of Hotelling's $T^2$ method that overlooks the changes and shifts in the distribution of latent features. Finally, we propose an adaptive weight-based Bayesian fusion (W-BF) framework to effectively aggregate the monitoring results from individual blocks. Our proposed W-BF framework automatically assigns higher weights to blocks based on how early anomalies occur in each block, thereby improving overall fault detection speed and accuracy. Among all these technical advancements, the key contributions of this work are:
\begin{enumerate}
    \item We introduce a novel autoencoder architecture OLAE to extract non-redundant and mutually independent dynamic features. Compared to existing autoencoder designs, OLAE demonstrates superior fault detection performance.
    \item We adopt a state-of-the-art quantile-based multivariate CUSUM framework to enable fast, accurate, and robust detection of process anomalies based on the mean shift in the distribution of extracted features. 
    \item We propose a novel W-BF approach to dynamically adjust the weights assigned to monitoring results from different blocks, which significantly enhances fault detection speed and accuracy.
\end{enumerate}

Dataset from the benchmark Tennessee Eastman Process (TEP) problem are used to evaluate our proposed MOLA framework. Some of the key capabilities of MOLA with respect to other related process monitoring methods are summarized in Table \ref{comparison}.

\begin{table}[H]\caption{A high-level comparison of key capabilities of different process monitoring methods.}\label{comparison}
\centering \vspace{1em}
\begin{adjustbox}{width=\columnwidth}
\begin{tabular}{ccccccccc}
\toprule
Capability                  & PCA & KPCA & DPCA & AE     & LSTM AE & Block PCA & Block LSTM AE & MOLA \\ \midrule
Cross-correlation       & \checkmark & \checkmark  & \checkmark  & \checkmark    & \checkmark     & \checkmark       & \checkmark           & \checkmark  \\
Dynamic                 & $\times$  & $\times$   & \checkmark  & $\times$     & \checkmark     & $\times$        & \checkmark           & \checkmark  \\
Nonlinearity            & $\times$  & \checkmark  & $\times$   & \checkmark    & \checkmark     & $\times$        & \checkmark           & \checkmark  \\
Orthogonality     & \checkmark & \checkmark  & \checkmark  & $\times$     & $\times$      & \checkmark       & $\times$            & \checkmark  \\
Large-scale monitoring& $\times$  & $\times$   & $\times$   & $\times$     & $\times$      & \checkmark       & \checkmark           & \checkmark  \\
Adaptive Weight         & $\times$  & $\times$   & $\times$   & $\times$     & $\times$      & $\times$        & $\times$            & \checkmark  \\
Distribution monitoring & $\times$  & $\times$   & $\times$   & $\times$     & $\times$      & $\times$        & $\times$            & \checkmark  \\ \bottomrule
\end{tabular}
\end{adjustbox}
\end{table}

The rest of this paper is organized as follows. Section 2 provides a brief review of LSTM and the quantile-based multivariate CUSUM method, followed by a detailed introduction to the proposed MOLA and the adaptive W-BF frameworks. Section 3 discusses the detailed steps involved in offline training and online monitoring of the proposed process monitoring framework. Next, in Section 4, we showcase the performance of our proposed methodology in the benchmark problem of the Tennessee Eastman Process (TEP), demonstrating its outstanding fault detection speed and accuracy compared to benchmark methods. To conclude, we summarize all results and learnings and discuss potential improvements for future research in Section 5.

\section{Preliminaries and Methods}
In this section, we provide the theoretical background of the backbone methods (e.g., LSTM and quantile-based multivariate CUSUM) upon which our proposed framework is based. We then formally introduce our proposed approaches, including OLAE and adaptive weight-based Bayesian fusion (W-BF).

\subsection{LSTM}
\begin{figure}[ht!]
    \centering
    \includegraphics[width=0.8\textwidth]{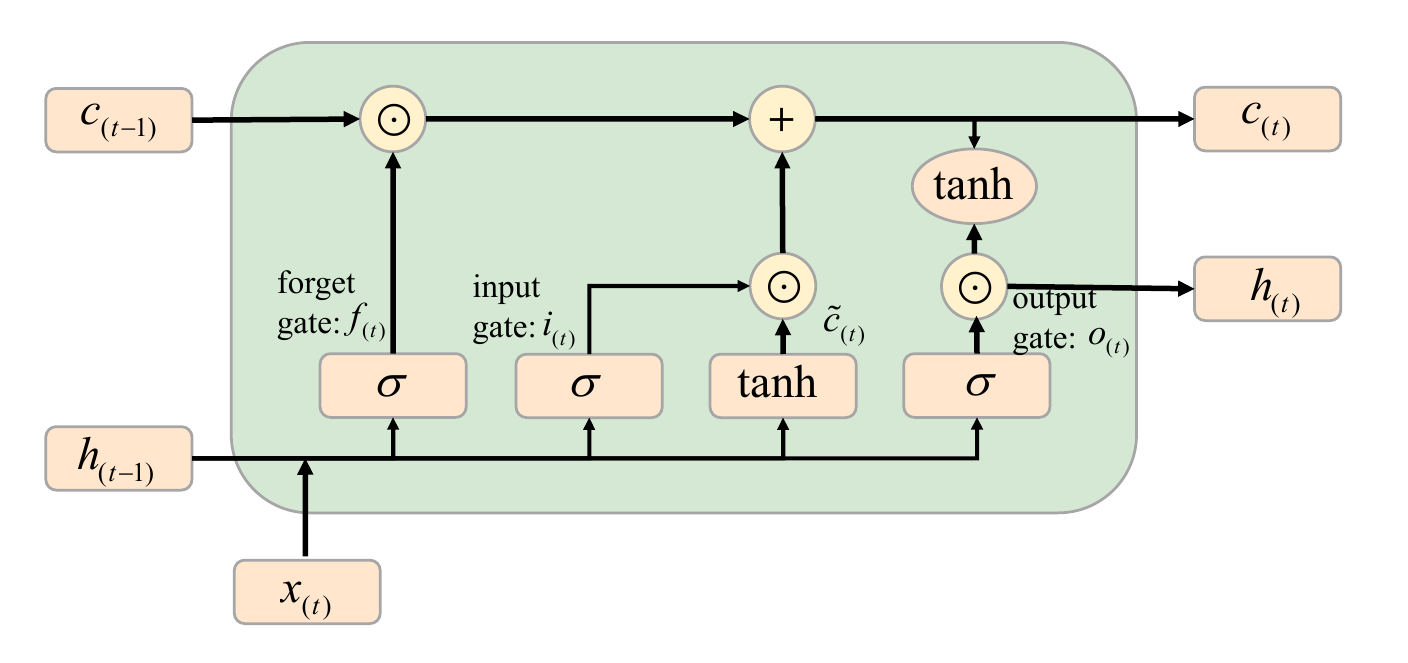}
    \vspace{-1em}
    \caption{Illustration of LSTM unit architecture featuring forget, input, and output gates.}
    \label{fig_LSTM}
\end{figure}

The recurrent neural network (RNN) is a class of neural network architectures specifically designed for time series modeling and prediction. This makes RNN-based methods particularly attractive in capturing the dynamic features of industrial process data \cite{zhang2019bidirectional}. However, traditional RNNs are prone to the issues of gradient explosion and gradient vanishing when working with long-term time series \cite{deng2023lstmed}. To overcome these issues, LSTM, an advanced RNN that uses ``gates'' to capture both long-term and short-term memory, is introduced, featuring a unique structural unit at its core \cite{ren2020batch}. As illustrated in Figure \ref{fig_LSTM}, an LSTM unit contains three crucial control gates: the forget gate $f(t)$, the input gate $i(t)$, and the output gate $o(t)$. These gates work collectively to ensure that essential information is consistently retained while the less important information is discarded. Therefore, LSTM achieves superior performance in capturing complex temporal dynamics embedded in chemical process systems. For more detailed information and mathematical descriptions about LSTM, readers are encouraged to refer to \cite{ma2022data}. 

\subsection{OLAE}
The LSTM AE is a specialized type of AE that seamlessly integrates LSTM with AE. This hybrid architecture enables efficient encoding and decoding of temporal sequences while capturing long-term dynamic features and dependencies within individual data streams at the same time. Similar to AE, the LSTM AE typically uses the reconstruction error as the loss function for model training. The Mean Squared Error (MSE) is one of the most widely adopted reconstruction error formulations:
\begin{equation}\label{eqn_mse}
\operatorname{Loss}_{\mathrm{MSE}}=\frac{1}{K} \sum_{i=1}^{K}(x_{i}-y_{i})^2,
\end{equation}
where $K$ is the total number of samples, $x_i$ represents the $i$-th original input vector, and $y_i$ is the corresponding reconstructed output vector by the LSTM AE. Nevertheless, as discussed earlier, the use of only Equation \eqref{eqn_mse} in the loss function can cause redundancies among the latent features, which will adversely affect the performance of LSTM AE in process monitoring tasks.

\begin{figure}[ht!]
    \centering
    \includegraphics[width=0.8\textwidth]{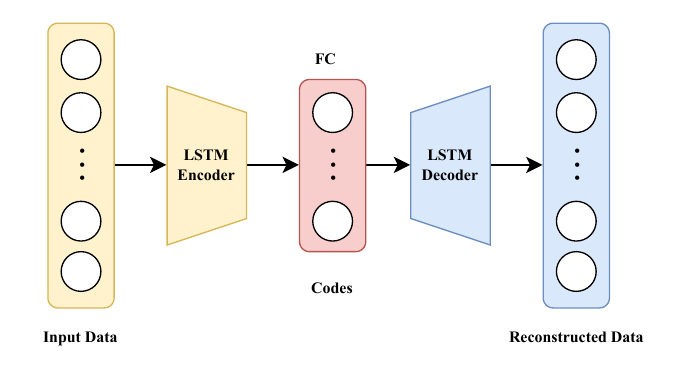}
    \vspace{-2em}
    \caption{Our proposed OLAE architecture consists of an LSTM encoder, an LSTM decoder, and a fully connected (FC) layer that leverages orthogonality. The FC layer is denoted as $C=\sigma(Wh+b)$, where $h$ and $C$ represent the output of the encoder and fully connected (FC) layer, respectively. Here, $b$ and $\sigma$ are the bias term and nonlinear activation function of the FC layer, respectively.} \label{fig_OLAE}
\end{figure}

Therefore, in the OLAE architecture as shown in Figure \ref{fig_OLAE}, we define an orthogonality-based loss function in Equation \eqref{eqn_ortho} to constrain the latent space output to generate non-redundant orthogonal latent features:
\begin{equation}\label{eqn_ortho}
\operatorname{Loss}_{\perp}=\left\|L_{0}(W)\right\|_{F}^{2} + \left |  \right | C^{T}C-I\left |  \right |_F^{2},\, L_{0}(W) = \left[\begin{array}{cccc}w_{1} w_{1}^{T} & w_{1} w_{2}^{T} & \dots & w_{1} w_{m}^{T} \\ w_{2} w_{1}^{T} & w_{2} w_{2}^{T} & \dots & w_{2} w_{m}^{T} \\\dots  & \dots & \dots  & \dots \\w_{m} w_{1}^{T} & w_{m} w_{2}^{T} & \dots & w_{m} w_{m}^{T}\end{array}\right],
\end{equation}
where $W = [w_1, w_2, \dots, w_m]^T$ is the weight matrix of the FC layer. The loss function $\operatorname{Loss}_{\perp}$ consists of two components. The primary purpose of having the first component is to drive the inner products between the weight vectors of neurons in the FC layer towards zero, which indicates that the linear projection features from the encoder layer to the FC layer are mutually orthogonal. This component can also be viewed as a regularization term, which drives the weights of the FC layer to smaller values during model training and thus effectively mitigates overfitting issues. Meanwhile, the second term ensures that the extracted latent features remain as mutually independent as possible even after undergoing nonlinear transformation ($\sigma$).

The overall loss function of OLAE is defined as:
\begin{equation}
\operatorname{Loss} = \operatorname{Loss}_{\mathrm{MSE}} + \operatorname{Loss}_{\perp}
\end{equation}

By minimizing $\operatorname{Loss}$, we ensure that the latent features are non-redundant and as mutually independent as possible.

\subsection{Monitoring statistics}

As previously discussed, AE-based process monitoring methods typically use the reconstruction error as the monitoring statistic for fault detection. This approach does not make full use of the lower-dimensional embeddings which could represent the underlying process dynamics. Thus, in this work, we propose to directly monitor the extracted features using the $T^2$ statistic defined as:
\begin{equation}
T^{2}= c \Lambda_{c}^{-1} c^{T}
\end{equation}
where \textit{c} represents the extracted codes, and $\Lambda_{c}$ is the covariance matrix of the codes. Based on the \textit{T}$^2$ statistic during in-control operations, the control limit for the monitoring statistic can be determined using the Kernel Density Estimation (KDE) method \cite{mao2018feature}.

While many process faults can be recognized based on changes in the numerical values of process variables being monitored, some faults are more represented by changes or variations in the distributions of process variables. To better detect the latter faults, in addition to employing the $T^2$ statistic, we adopt the quantile-based multivariate CUSUM method recently developed by Ye and Liu \cite{ye2022generic}. The basic idea behind this new CUSUM method is to detect process anomalies by monitoring any mean shifts in data stream distributions. Compared to traditional multivariate CUSUM techniques, this novel framework handles nonparametric and heterogeneous data streams for the first time. Previously, Jiang successfully applied this method to chemical process monitoring and achieved promising results\cite{jiang2023online}. Here, we build upon this CUSUM framework to monitor any subtle deviations in the distribution of dynamic orthogonal latent features.

For each process variable $x=1,\dots,p$, the data collected under normal operating conditions can be divided into $d$ quantiles: $I_{x,1}=(-\infty,q_{x,1}],\, I_{x,2}=(q_{x,1},q_{x,2}],\, \dots,\, I_{x,d}=(q_{x,d-1}, +\infty)$, such that each quantile contains exactly $\frac{1}{d}$ of the in-control data samples. Next, we define a vector $Y_x(t)=(Y_{x,1}(t),Y_{x,2}(t),\dots,Y_{x,d}(t))^T$ for each data stream $x$ at time $t$, where $Y_{x,l} = \mathbb{I}$\{$G_x(t) \in I_{x,l}$\} with $l=1,\dots,d$. Here, $\mathbb{I}$\{$G_x(t)\in I_{x,l}\}$ is an indicator function that equals 1 when the online measurement $G_x(t)$ lies in the interval $I_{x,l}$ and 0 otherwise. Then, we define two vectors $A_{x}^{+}(t)=[A_{x,1}^{+}(t),\dots,A_{x,d-1}^{+}(t)]^T$ and $A_{x}^{-}(t)=[A_{x,1}^{-}(t),\dots,A_{x,d-1}^{-}(t)]^T$, where $A_{x,l}^{+}(t)= 1- \sum_{i=1}^{l} Y_{x,i}(t)$, $A_{x,l}^{-}(t) = \sum_{i=1}^{l} Y_{x,i}(t)$. Ye and Liu \cite{ye2022generic} showed that detecting the mean shifts in the distribution of $G_x(t)$ is equivalent to detecting shifts in the distribution of $A^{+}_{x,l}(t)$ and $A^{-}_{x,l}(t)$ with respect to their expected values, which are $1-\frac{l}{d}$ and $\frac{l}{d}$, respectively. With this, the multivariate CUSUM procedure originally proposed by Qiu and Hawkins \cite{qiu2003nonparametric} can now be employed to detect these the mean shifts of $A^{\pm}_{x,l}(t)$. This is done by defining a variable $C_x^{\pm}(t)$ as:
\begin{equation}
    \begin{aligned}
        C_x^{\pm }(t) = & \left[A_{x}^{\pm}(t) - \mathbb{E}(A^\pm_x(t)) + S_{x}^{\pm 0}(t-1)-S_{x}^{\pm 1}(t-1) \right]^{T} \\
        & \cdot [\diag(\mathbb{E}(A^\pm_x(t)) + S_{x}^{\pm 1}(t-1))]^{-1}\\
        & \cdot \left[S_{x}^{\pm 0}(t-1) + S_{x}^{\pm 1}(t-1) - \mathbb{E}(A^\pm_x(t)) + A_{x}^{\pm}(t)\right],
    \end{aligned}
\end{equation}
where $S_{x}^{\pm 0}(t)$ and $S_{x}^{\pm 1}(t)$ are $(d-1)$-dimensional vectors defined as follows:
\begin{equation}\label{eqn_S}
    \begin{cases}
    S_{x}^{\pm 0}(t)=0,\, S_{x}^{\pm 1}(t) = 0 \qquad \text{ if }\ C_{x}^{\pm}(t) \leq k;\\
    S_{x}^{\pm 0}(t) = \frac{(S_{x}^{\pm 0}(t-1) + A_{x}^{\pm}(t))(C_{x}^{\pm}(t) - k)}{C_{x}^{\pm}(t)};\\
    S_{x}^{\pm 1}(t) = \frac{(S_{x}^{\pm 1}(t-1) + \mathbb{E}(A^\pm_x(t)))(C_{x}^{\pm}(t) - k)}{C_{x}^{\pm}(t)} \qquad \text{ if } \ C_{x}^{\pm}(t) > k.
    \end{cases}
\end{equation}

In Equation \eqref{eqn_S}, $k$ is a pre-computed allowance parameter that restarts the CUSUM procedure by resetting the local statistic back to 0 if there is no evidence of upward or downward mean shift after a while \cite{ye2022generic}. The local statistic $W_x^{\pm}(t)$ for detecting any mean shift in the upward ($+$) or downward ($-$) direction is calculated for each time $t$ as:
\begin{equation}
W_{x}^{\pm }(t) = \max(0,C_{x}^\pm(t) -k)
\end{equation}

Overall, we monitor the two-sided statistic $W_{x}(t)=\max(W_{x}^-(t),W_{x}^+(t))$ for both upward and downward mean shifts. An alarm is raised (i.e., an anomaly is detected) when the monitoring statistic, $\sum_{(x)=1}^{r} W_{(x)}(t)$, defined as the sum of the largest $r$ local statistics $W_{x}$ at each time $t$, exceeds a threshold $h$ that is related to the pre-specified false alarm rate (e.g., 0.0027 for the typical $3\sigma$-limit) \cite{mei}. The corresponding stopping time $T$ is thus:
\begin{equation}\label{eqn_stoppingtime}
T = \inf \left\{t>0: \sum_{(x)=1}^{r} W_{(x)}(t) \geq h  \right\}.
\end{equation}

More information about the theory and application of quantile-based multivariate CUSUM, including detailed derivations of the mathematical formulations above, can be found in Ye and Liu \cite{ye2022generic}.

\subsection{Adaptive weight-based Bayesian fusion strategy}

Here, we describe our multi-block monitoring framework for large-scale, complex industrial systems. Each block is monitored by two metrics (resp. $T^2$ and $W$) using two approaches (resp. Hotelling's $T^2$ and quantile-based CUSUM). We adopt a Bayesian data fusion framework to aggregate the two monitoring results into a single monitoring metric in each block, as well as to aggregate all block-level metrics into a single plant-wide fault index (PFI). Bayesian fusion has shown remarkable robustness and capabilities in integrating information from diverse sources. It leverages prior knowledge and real-time measurements to compute posterior probabilities using Bayes' theorem. In this work, we extend the classic Bayesian fusion methodology and propose an adaptive weight-based Bayesian fusion (W-BF) method. The idea is to dynamically adjust fusion weights based on the relative ranking of the current monitoring statistics from each block. Under this framework, the probability of sample $X_i^n(t)$ of block $n$ and monitoring metric $i$ in normal and fault conditions are given by:
\begin{equation}
\mathbb{P}_i^n(X_{i}^n(t)|N) = \exp\left(-\frac{S_{i}^n(t)}{S_{i,\lim}^n}\right);\; \mathbb{P}_i^n(X_{i}^n(t)|F) = \exp \left( -\frac{S_{i,\lim}^n}{S_{i}^n(t)}\right),
\end{equation}
where $S_{i}^n(t)$ and $S_{I,\lim}^n$ denote the current value and control limit of the monitoring metric $i$, respectively. With this, the posterior probability can be calculated using Bayes' rule as:
\begin{equation}
\begin{aligned}
&\mathbb{P}_i^n(F|X_{i}^n(t))=\frac{\mathbb{P}_i^n(X_{i}^n(t)|F)\mathbb{P}_i^n(F)}{\mathbb{P}_i^n(X_{i}^n(t)|F)\mathbb{P}_i^n(F)+\mathbb{P}_i^n(X_{i}^n(t)|N)\mathbb{P}_i^n(N)}; \\
    &\mathbb{P}_i^n(N)=1-\alpha; \\
    &\mathbb{P}_i^n(F)=\alpha, 
\end{aligned}
\end{equation}
where $\mathbb{P}_i^n(N)$  and $\mathbb{P}_i^n(F)$ denote prior probabilities under normal and abnormal conditions, respectively; $\alpha$ is the significance level, which is taken to be 0.01\cite{rong2021multi,ge2015plant}. 

Thus, the fused monitoring statistic of block $n$, $B^n(t)$, is determined as:
\begin{equation}
B^n(t)=\frac{\sum_{i=1}^{2}w_{i}^n(t)\mathbb{P}_i^n(X_{i}^n(t)|F)\mathbb{P}^n_{i}(F|X_{i}^n(t))}{ \sum_{i=1}^{2} w_{i}^n(t)\mathbb{P}_i^n(X_{i}^n(t)|F)},
\end{equation}
 where $w_{i}^n(t)$ represents the weight for monitoring metric $i$, which is dynamically updated at every time $t$ as follows:
\begin{equation}
w_{i}^n(t)=\frac{\exp((S_{i}^n(t)-S_{i,\lim}^n)/S_{i,\lim}^n)}{\sum_{1}^{2} \exp((S_{i}^n(t)-S_{i,\lim}^n)/S_{i,\lim}^n)} 
\end{equation}

Following a similar procedure, we adopt adaptive W-BF once again to aggregate $B^n(t)$ of all blocks to obtain the PFI. First, we derive the following block-wise probabilities:
\begin{equation}
\begin{aligned}
&\mathbb{P}^n(B^n(t)|N) = \exp\left(-\frac{B^n(t)}{B^n_{\lim}}\right);\; \mathbb{P}^n(B^n(t)|F) = \exp \left( -\frac{B^n_{\lim}}{B^n(t)}\right); \\
&\mathbb{P}^n(F|B^n(t))=\frac{\mathbb{P}^n(B^n(t)|F)\mathbb{P}^n(F)}{\mathbb{P}^n(B^n(t)|F)\mathbb{P}^n(F)+\mathbb{P}^n(B^n(t)|N)\mathbb{P}^n(N)}; \\
&\mathbb{P}^n(N)=1-\alpha; \\
&\mathbb{P}^n(F)=\alpha, 
\end{aligned}
\end{equation}
where $\mathbb{P}^n(B^n(t)|N)$ and $\mathbb{P}^n(B^n(t)|F)$ represent the probability of block $n$ in normal and fault conditions at time t, respectively; $\mathbb{P}^n(F|B^n(t))$, $\mathbb{P}^n(N)$ and $\mathbb{P}^n(F)$ are the posterior probability, the prior integration probability and prior failure probability for block $n$, respectively. Finally, the PFI is calculated as:
\begin{equation} \begin{aligned}
&\operatorname{PFI}(t)=\frac{\sum_{n=1}^{N}w_n(t)\mathbb{P}^n(B^n(t)|F)\mathbb{P}^n(F|B^n(t))}{ {\textstyle \sum_{n=1}^{N}} w_n(t)\mathbb{P}^n(B^n(t)|F)};\\
    &w_{n}(t)=\frac{\exp((B^n(t)-B^n_{\lim})/B^n_{\lim})}{\sum_{n=1}^{N} \exp((B^n(t)-B^n_{\lim})/B^n_{\lim})}, 
\end{aligned}
\end{equation}
where $N$ is the number of blocks and $B^n_{\lim}$ is the control limit of the fused monitoring statistic of block $n$.

\section{The MOLA Fault Dection Framework}

\begin{figure}[ht!]
    \centering
    \includegraphics[width=\textwidth]{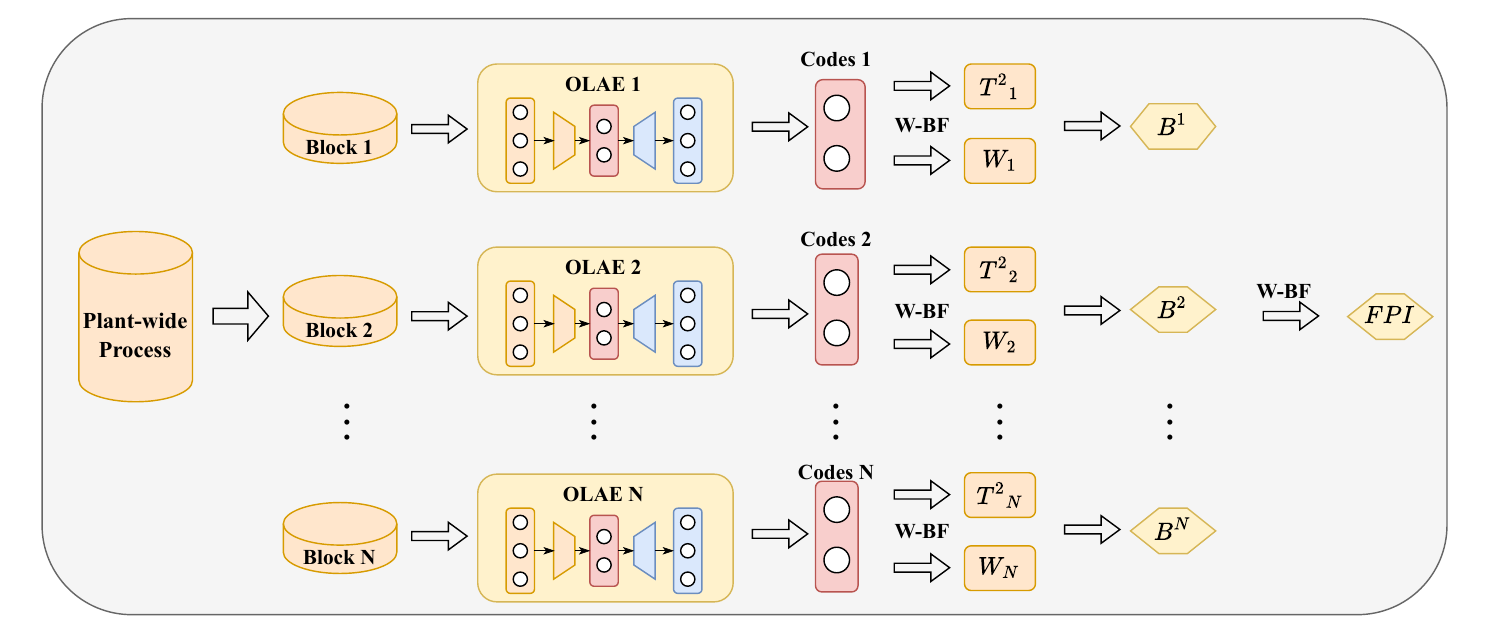}
    \vspace{-2em}
    \caption{The MOLA process monitoring framework features an OLAE model for each block and adaptive W-BF for data fusion.} \label{fig_MOLA}
\end{figure}

Now that all components of our proposed process monitoring framework have been introduced, we will move on to discuss how these components are integrated with our MOLA framework during offline learning and online monitoring stages. The overall flowchart of MOLA is shown in Figure \ref{fig_MOLA}.

The steps involved in the offline learning stage are outlined below:
\begin{enumerate}[left=1pt,nosep,label=\textbf{Step \arabic*:}]
    \item Historical in-control data are collected, normalized, and standardized.
    \item Based on process knowledge, process variables are divided into blocks. For each block, we establish a local OLAE model. We use 70\% of the historical in-control data to build the OLAE model, whereas the remaining data serve as the validation set to determine the optimal hyperparameters for the model.
    \item The validation data are sent to the OLAE model of each block to obtain the corresponding latent features or codes. Then, we calculate $T^2$ and $W$ for each block and determine their control limits.
    \item For each block, we implement the adaptive W-BF strategy to obtain a fused monitoring statistic.
    \item Based on the fused monitoring statistics of each block, we apply the adaptive W-BF technique again to determine the PFI for the overall process.
\end{enumerate}

The steps involved in the online monitoring stage are:
\begin{enumerate}[left=1pt,nosep,label=\textbf{Step \arabic*:}]
    \item As online data are collected, they are standardized based on the mean and variance of the in-control data collected during the offline learning phase.
    \item Following the block assignments, standardized online data are sent to their corresponding block's OLAE model to obtain the codes and monitoring statistics $T^2$ and $W$.
    \item The fused monitoring statistic is calculated following the adaptive W-BF strategy for each block.
    \item Using the adaptive W-BF method again, we obtain the process-level PFI from the local statistics of all blocks. If the PFI is greater than our pre-defined significance level $\alpha$ (0.01), a fault is declared; otherwise, the process is under normal operation and the monitoring continues.
\end{enumerate}

\section{Case Study}

\begin{figure}[ht!]
    \centering
    \includegraphics[width=\textwidth]{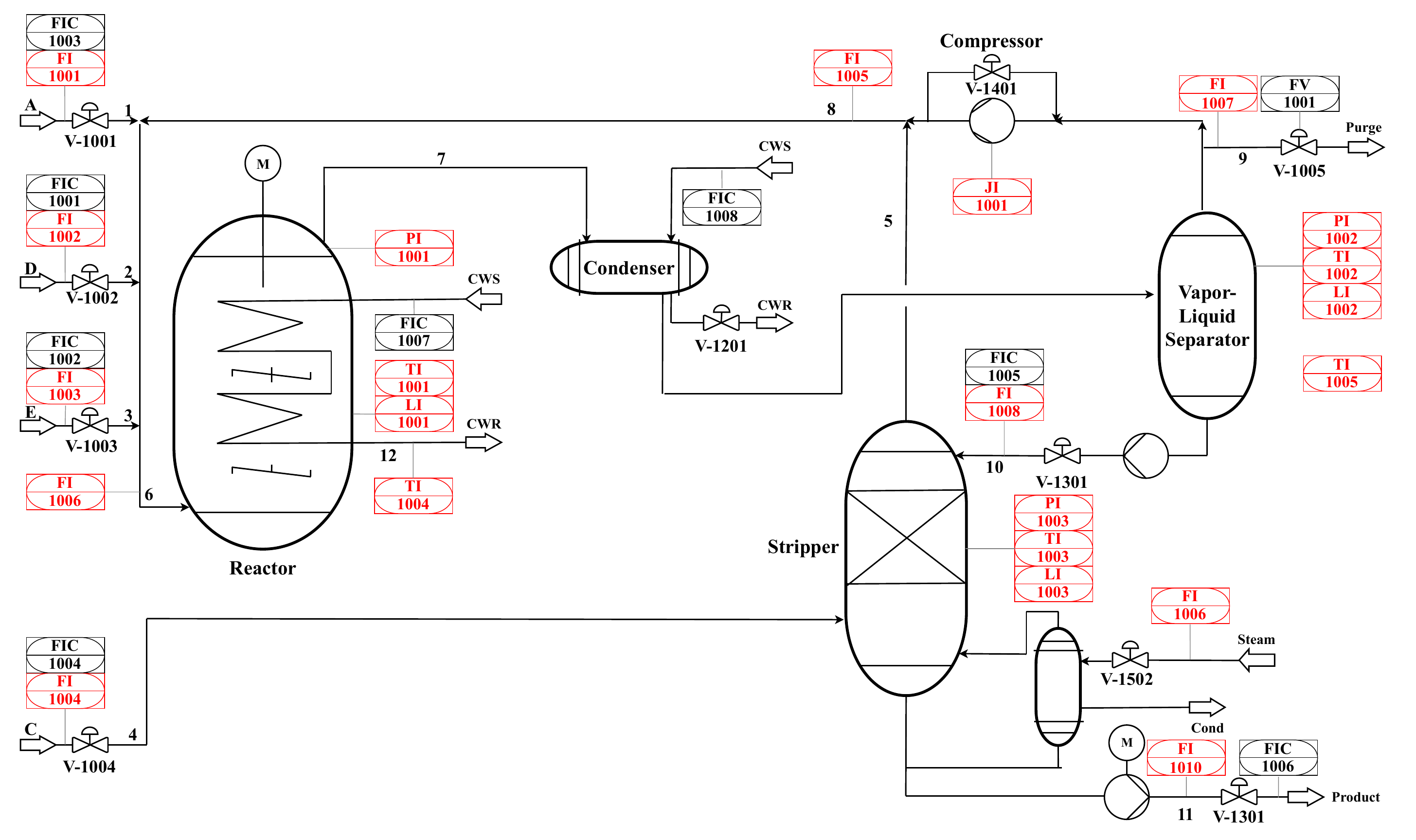}
    \vspace{-2em}
    \caption{Process flow diagram of the TEP.}
    \label{fig_TEP}
\end{figure}

The Tennessee Eastman Process (TEP) is a simulation process developed by the Eastman Chemical Company based on an actual chemical process \cite{downs1993plant}. The TEP has been widely adopted as a benchmark for chemical process control, optimization, and monitoring. As illustrated in the process flow diagram of Figure \ref{fig_TEP}, the TEP contains five major unit operations, which are associated with 12 manipulated variables and 41 measured variables in total. Among them, 31 variables (listed in Table \ref{table_TEPvariables}) are typically selected to conduct process monitoring, as the remaining variables have relatively low sampling frequencies. In multi-block process monitoring methods, the choice of block division and assignment approach directly influences the process monitoring performance. Existing block division methods include process knowledge-based methods, variable relation-based methods, and fault information-aided methods \cite{jiang2019review}. In this work, we adopt the process knowledge-based approach developed by Zhu et al. \cite{zhu2017distributed} and divide the entire process into four distinct blocks ($N = 4$) as outlined in Table \ref{table_TEPblock}. The basic idea is to assign process variables associated with the same equipment into a single block. And due to the relatively small number of process variables being measured for the condenser and compressor, we assign these variables to the nearest separator block.

In this work, the simulation data are generated from a Simulink implementation with a sampling frequency of one minute \cite{bathelt2015revision}. Initially, a dataset comprising 60,000 samples under normal operating conditions is generated and used for offline learning. Subsequently, an additional 500 datasets, each containing 2,400 samples, are generated to determine the control limits in process monitoring models. Furthermore, this Simulink implementation can simulate 20 types of process faults (see Table \ref{table_TEPfaults}) as test datasets for process online monitoring. It is worth noting that any process fault is introduced at the 600th sample in each test dataset.

\begin{table}[H]\caption{The detailed information of the monitoring variables in the TEP.}\label{table_TEPvariables}
\centering \vspace{1em}
\begin{adjustbox}{width=\columnwidth}
\begin{tabular}{ccc||ccc}
\toprule
\textbf{No.} & \textbf{Variable} & \textbf{Description}                    & \textbf{No.} & \textbf{Variable} & \textbf{Description}                       \\ \midrule
\textbf{0}   & FI-1001  & A feed (stream 1)                       & \textbf{16}          & FI-1009              & Stripper underflow (stream 11)             \\
\textbf{1}   & FI-1002  & D feed (stream 2)                       & \textbf{17}          & TI-1003              & Stripper temperature                       \\
\textbf{2}   & FI-1003  & E feed (stream 3)                       & \textbf{18}          & FI-1010              & Stripper steam flow                        \\
\textbf{3}   & FI-1004  & A and C feed (stream 4)                 & \textbf{19}          & JI-1001              & Compressor work                            \\
\textbf{4}   & FI-1005  & Recycle flow (stream 8)                 & \textbf{20}          & TI-1004              & Reactor cooling water outlet temperature   \\
\textbf{5}   & FI-1006  & Reactor feed rate (stream 6)            & \textbf{21}          & TI-1005              & Separator cooling water outlet temperature \\
\textbf{6}   & PI-1001  & Reactor pressure                        & \textbf{22}          & FIC-1001             & D feed flow (stream 2)                     \\
\textbf{7}   & LI-1001  & Reactor level                           & \textbf{23}          & FIC-1002             & E feed flow (stream 3)                     \\
\textbf{8}   & TI-1001  & Reactor temperature                     & \textbf{24}          & FIC-1003             & A feed flow (stream 1)                     \\
\textbf{9}   & FI-1007  & Purge rate (stream 9)                   & \textbf{25}          & FIC-1004             & A and C feed flow (stream 4)               \\
\textbf{10}  & TI-1002  & Product separator temperature           & \textbf{26}          & FV-1001              & Purge valve (stream 9)                     \\
\textbf{11}  & LI-1002  & Product separator level                 & \textbf{27}          & FIC-1005             & Separator pot liquid flow (stream 10)      \\
\textbf{12}  & PI-1002  & Product separator pressure              & \textbf{28}          & FIC-1006             & Stripper liquid prod flow (stream 11)      \\
\textbf{13}  & FI-1008  & Product separator underflow (stream 10) & \textbf{29}          & FIC-1007             & Reactor cooling water flow                 \\
\textbf{14}  & LI-1003  & Stripper level                          & \textbf{30}          & FIC-1008             & Condenser cooling water flow               \\
\textbf{15}  & PI-1003  & Stripper pressure                       & \multicolumn{1}{l}{} & \multicolumn{1}{l}{} & \multicolumn{1}{l}{}                       \\ \bottomrule
\end{tabular}
\end{adjustbox}
\end{table}

\begin{table}[H]\caption{Divided blocks of the monitoring variables.}\label{table_TEPblock}
\centering \vspace{1em}
\begin{tabular}{ccc}
\toprule
\textbf{Block} & \textbf{Variables}           & \textbf{Description}                     \\ \midrule
\textbf{1}     & 0,1,2,4,5,22,23,24           & Input                                    \\
\textbf{2}     & 6,7,8,20,29                  & Reactor                                  \\
\textbf{3}     & 9,10,11,12,13,19,21,26,27,30 & Separator, compressor and condenser \\
\textbf{4}     & 3,14,15,16,17,18,25,28       & Stripper                                 \\ \bottomrule
\end{tabular}
\end{table}

\begin{table}[H] \caption{The detailed information of the TEP faults. \label{table_TEPfaults}}
\centering \vspace{1em}
\begin{tabular}{ccc}
\toprule
\textbf{Fault No.} & \textbf{Process variable}                              & \textbf{Type}    \\ \midrule
\textbf{1}         & A/C feed ratio, B composition constant (stream 4)      & Step             \\
\textbf{2}         & B composition, A/C ratio constant (stream 4)           & Step             \\
\textbf{3}         & D feed temperature (stream 2)                          & Step             \\
\textbf{4}         & Reactor cooling water inlet temperature                & Step             \\
\textbf{5}         & Condenser cooling water inlet temperature              & Step             \\
\textbf{6}         & A feed loss (stream 1)                                 & Step             \\
\textbf{7}         & C header pressure loss-reduced availability (stream 4) & Step             \\
\textbf{8}         & A, B, C feed composition (stream 4)                    & Random variation \\
\textbf{9}         & D feed temperature (stream 2)                          & Random variation \\
\textbf{10}        & C feed temperature (stream 4)                          & Random variation \\
\textbf{11}        & Reactor cooling water inlet temperature                & Random variation \\
\textbf{12}        & Condenser cooling water inlet temperature              & Random variation \\
\textbf{13}        & Reaction kinetics                                      & Slow drift       \\
\textbf{14}        & Reactor cooling water valve                            & Sticking         \\
\textbf{15}        & Condenser cooling water valve                          & Sticking         \\
\textbf{16}        & Unknown                                                & Unknown          \\
\textbf{17}        & Unknown                                                & Unknown          \\
\textbf{18}        & Unknown                                                & Unknown          \\
\textbf{19}        & Unknown                                                & Unknown          \\
\textbf{20}        & Unknown                                                & Unknown          \\ \bottomrule
\end{tabular}
\end{table}

We compare our proposed framework with other process monitoring models, including PCA, AE, LSTM AE, Block PCA, and Block LSTM AE. The complete structure of each neural network implemented in our framework is listed in Table \ref{structure}. In this work, all neural networks use the Rectified Linear Unit (ReLU) activation function. The number of iterations is determined using early stopping criteria. Specifically, we stop the training process when the loss function value on the validation dataset does not decrease for consecutive 30 iterations. The detailed structure for different neural networks is illustrated in Table \ref{structure}. Tables \ref{tab4} and \ref{tab5} show the performance of these models in terms of Fault Detection Delay (FDD) and Fault Detection Rate (FDR), respectively. It can be seen that, in general, the incorporation of a multi-block monitoring strategy not only reduces the FDD but also increases the FDR of original process monitoring methods. This validates the effectiveness of our proposed multi-block method and data fusion technique. In particular, our MOLA framework exhibits superior performance over all other frameworks in terms of FDR and FDD. For the vast majority of fault cases, particularly faults 13, 15, 16, 24, and 18, MOLA can detect faults tens to hundreds of minutes ahead of other methods, providing invaluable time buffers for engineers and operators to take appropriate control actions. For a few faults, such as faults 3, 11, 14, and 20, MOLA falls behind the best-performing method by just a few minutes. However, considering that the differences are small and that MOLA achieves significant improvements in FDR in these faults, MOLA still outperforms other methods by a considerable margin. For example, for faults 3, 5, 9, 15, and 16, while other methods perform poorly on these faults, MOLA increases the FRD by 4.6 to 30 times. Note that faults 3, 9, and 15 are known to be notoriously difficult to detect due to their intricate process dynamics. Meanwhile, the fault detection rates for MOLA on these three faults range between 88.3 and 98.5\%, demonstrating MOLA's exceptional capabilities in tackling challenging fault detection scenarios.

\begin{table}
\caption{Detailed structure for different neural networks.\label{structure}}
\centering
\begin{tabular}{ccc}
\toprule
Model & Structure & Activation function \\
\midrule
AE & FC(64)-FC(16)-FC(64) & ReLU \\
LSTM AE & LSTM(64)-FC(16)-LSTM(64) & ReLU \\
Block LSTM AE & LSTM(15)-FC(5)-LSTM(15) & ReLU \\
OLAE & LSTM(15)-FC(5)-LSTM(15) & ReLU \\
\bottomrule

\end{tabular}

\end{table}

\begin{figure}[H]
\centering
\includegraphics[width=\textwidth]{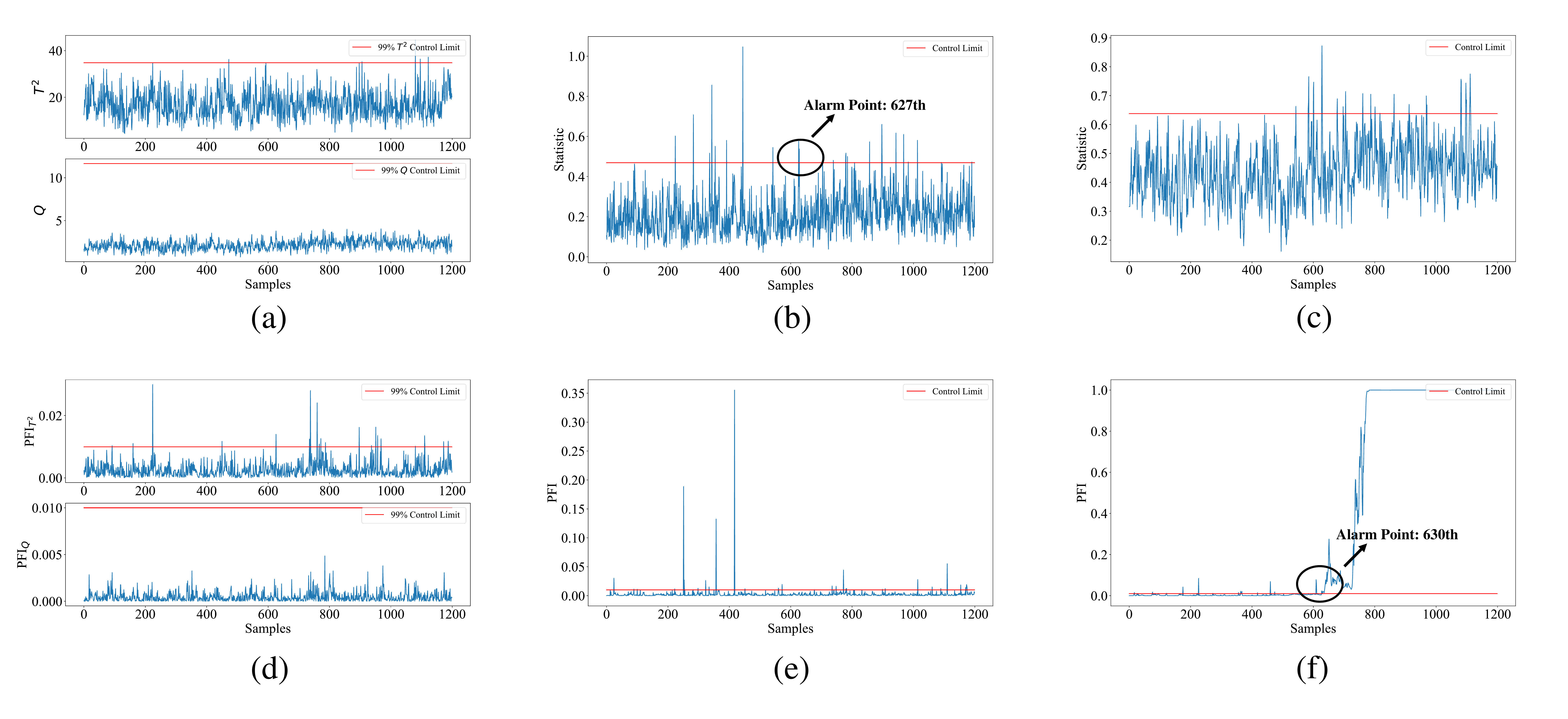}
\vspace{-2em}
\caption{The process monitoring results of Fault 3 based on (a) PCA, (b) AE, (c) LSTM AE, (d) Block PCA, (e) Block LSTM AE and (f) MOLA.\label{fig_fault3}}
\end{figure}

We now take a close look at fault 3, whose process monitoring results for various methods are summarized in Figure \ref{fig_fault3}. As we can see, while the LSTM AE-based approach successfully detects the fault at the 627th sample (i.e., 27 samples after the process fault is introduced into the simulation), its monitoring statistic only briefly exceeds the control limit and fails to raise a continuous alarm. Similarly, most other methods are unable to raise any valid alarm for this fault, as in practice alarms would only be considered effective when the monitoring statistic exceeds the control limit multiple consecutive times (e.g., three)\cite{alakent2024early}. If a process monitoring model fails to continuously report faults, plant operators may erroneously assume that the alarm is false or the process has returned back to normal. On the other hand, our MOLA framework captures the fault starting at the 630th sample, and subsequently raises alarms continuously, making it a useful process monitoring framework in practice.

\begin{figure}[H]
\centering
\includegraphics[width=\textwidth]{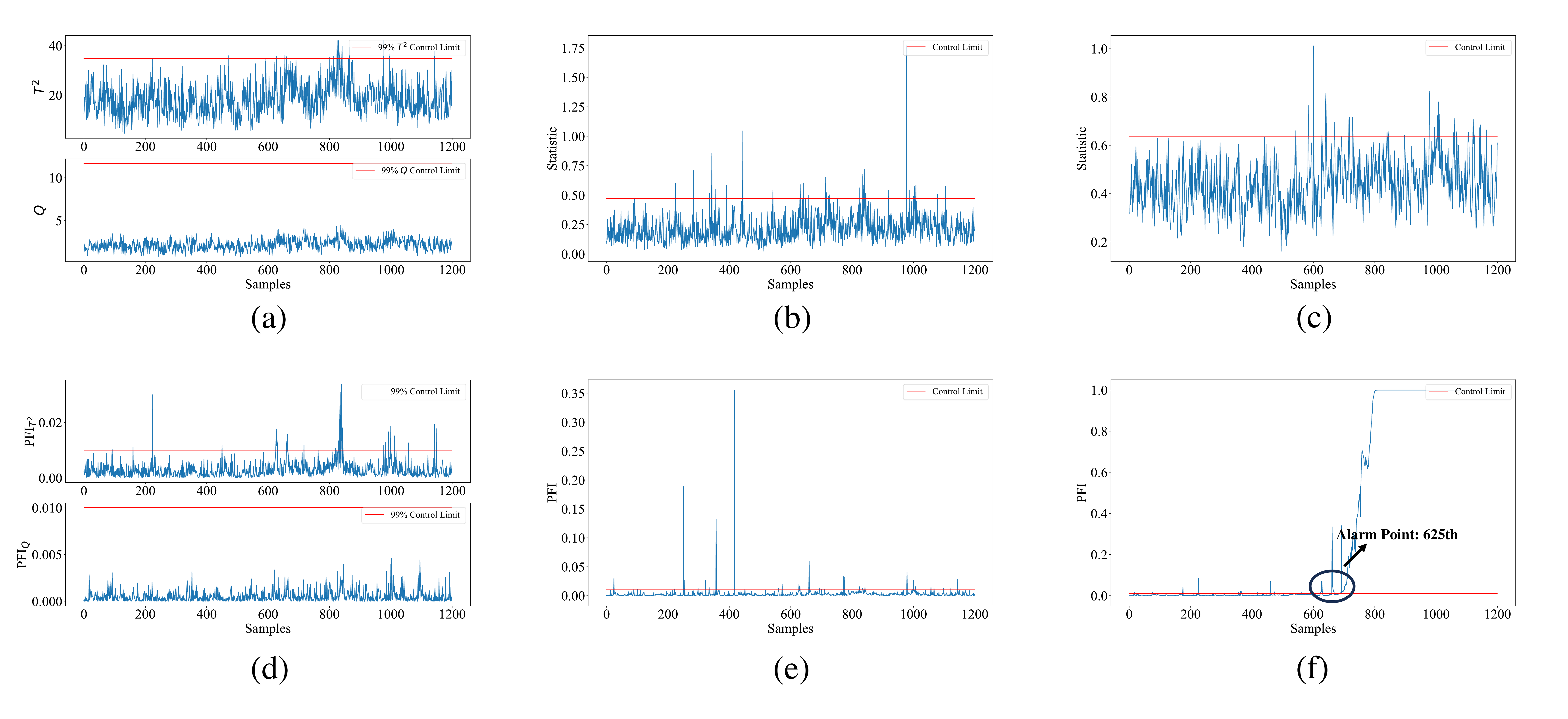}
\vspace{-2em}
\caption{The process monitoring results of Fault 9 based on (a) PCA, (b) AE, (c) LSTM AE, (d) Block PCA, (e) Block LSTM AE and (f) MOLA.\label{fig_fault9}}
\end{figure}

Similarly, for fault 9 (see Figure \ref{fig_fault9}), the MOLA framework successfully raises the alarm at the 625th sample and sustains the alarm, whereas all other methods remain ineffective in raising any meaningful alarm. Although MOLA occasionally produces false alarms during fault-free periods (see Table \ref{tab6} for the false alarm rates), these false alarms are isolated incidents of brief statistical excursions at a single sample, which have minimal impact in practice. These false alarms are typically attributed to suboptimal tuning of hyperparameters and can be resolved when more comprehensive model training is conducted.

\begin{table}[H]\caption{Comparison of FDD results in the TEP.\label{tab4}}
\centering \vspace{1em}
\begin{adjustbox}{width=\columnwidth}
\begin{tabular}{ccccccccc}
\toprule
\textbf{Fault No.} & \textbf{\begin{tabular}[c]{@{}c@{}}PCA\\ ($T^2$)\end{tabular}} & \textbf{\begin{tabular}[c]{@{}c@{}}PCA\\ (Q)\end{tabular}} & \textbf{AE} & \textbf{LSTM AE} & \textbf{\begin{tabular}[c]{@{}c@{}}Block PCA\\ ($T^2$)\end{tabular}} & \textbf{\begin{tabular}[c]{@{}c@{}}Block PCA\\ (Q)\end{tabular}} & \textbf{\begin{tabular}[c]{@{}c@{}}Block \\ LSTM AE\end{tabular}} & \textbf{MOLA} \\ \midrule
\textbf{1}   & 3   & 18   & 3  & 1  & 1  & 15 & 4  & \textbf{1}    \\
\textbf{2}   & 25  & 140  & 22 & 20 & 19 & 726  & 18 & \textbf{16}   \\
\textbf{3}   &--                                                             &--                                                         &--          & \textbf{27}      &--&--& 572                                                               & 30            \\
\textbf{4}   & 0                                                              & 0                                                          & 0           & 0                & 0                                                                    & 0                                                                & 0                                                                 & \textbf{0}    \\
\textbf{5}   &--                                                             &--                                                         &--          & 208              & 2                                                                    &--                                                               & 3                                                                 & \textbf{2}    \\
\textbf{6}   & 0                                                              & 0                                                          & 0           & 0                & 0                                                                    & 0                                                                & 0                                                                 & \textbf{0}    \\
\textbf{7}   & 0                                                              & 0                                                          & 0           & 0                & 0                                                                    & 0                                                                & 0                                                                 & \textbf{0}    \\
\textbf{8}   & 29                                                             & 200                                                        & 26          & 28               & 25                                                                   & 45                                                               & 27                                                                & \textbf{25}   \\
\textbf{9}   & 229                                                            &--                                                         &--          & 40               & 26                                                                   &--                                                               & 233                                                               & \textbf{25}   \\
\textbf{10}  & 91                                                             & 90                                                         & 77          & 77               & 74                                                                   & 188                                                              & 74                                                                & \textbf{70}   \\
\textbf{11}  & 36                                                             & 41                                                         & 36          & \textbf{22}      & 36                                                                   & 37                                                               & 36                                                                & 36            \\
\textbf{12}  & 66                                                             &--                                                         & 65          & 63               & 66                                                                   & 475                                                              & \textbf{48}                                                       & 60            \\
\textbf{13}  & 94                                                             & 317                                                        & 86          & 86               & 89                                                                   & 307                                                              & 91                                                                & \textbf{9}    \\
\textbf{14}  &--                                                             & 7                                                          & 3           & \textbf{0}       & 2                                                                    & 3                                                                & 2                                                                 & 2             \\
\textbf{15}  & 191                                                            &--                                                         &--          &--               &--                                                                   &--                                                               & 191                                                               & \textbf{6}    \\
\textbf{16}  &--                                                             &--                                                         &--          &--               &--                                                                   &--                                                               &--                                                                & \textbf{21}   \\
\textbf{17}  & 57                                                             & 62                                                         & 55          & 55               & 56                                                                   & 58                                                               & 57                                                                & \textbf{24}   \\
\textbf{18}  & 267                                                            & 280                                                        & 242         & 241              & 259                                                                  & 259                                                              & 268                                                               & \textbf{147}  \\
\textbf{19}  & 30                                                             & 32                                                         & 15          & 16               & 21                                                                   & 32                                                               & 12                                                                & \textbf{11}   \\
\textbf{20}  & 141                                                            & 178                                                        & 129         & 128              & 138                                                                  & 164                                                              & \textbf{123}                                                      & 124           \\ \bottomrule
\end{tabular}
\end{adjustbox}
\noindent{\footnotesize{* ``0'' indicates that the fault is detected at the time of introduction, while ``--'' indicates that the fault is not effectively detected.}}
\end{table}

\begin{table}[H]\caption{Comparison of FDR results in TEP. Here, Q is the square prediction error, and \% Improvement measures the \% difference in FDR between MOLA and the best-performing method. \label{tab5}}
\centering \vspace{0.5em}
\begin{adjustbox}{width=\columnwidth}
\begin{tabular}{cccccccccc}
\toprule
No.         & \textbf{\begin{tabular}[c]{@{}c@{}}PCA\\ ($T^2$)\end{tabular}} & \textbf{\begin{tabular}[c]{@{}c@{}}PCA\\ (Q)\end{tabular}} & \textbf{AE} & \textbf{LSTM AE} & \textbf{\begin{tabular}[c]{@{}c@{}}Block PCA\\ ($T^2$)\end{tabular}} & \textbf{\begin{tabular}[c]{@{}c@{}}Block PCA\\ (Q)\end{tabular}} & \textbf{\begin{tabular}[c]{@{}c@{}}Block \\ LSTM AE\end{tabular}} & \textbf{MOLA} & \textbf{\% Improvement} \\ \midrule
\textbf{1}  & 0.9967                                                         & 0.9700                                                     & 0.9950      & 0.9967           & 0.9983                                                               & 0.9750                                                           & 0.9950                                                            & \textbf{0.9983}& 0.00                \\
\textbf{2}  & 0.9617                                                         & 0.6850                                                     & 0.9667      & 0.9650           & 0.9683                                                               & 0.7200                                                           & 0.9700                                                            & \textbf{0.9767}& 0.68                \\
\textbf{3}  & 0.0067                                                         & 0.0000                                                     & 0.0183      & 0.0467           & 0.0267                                                               & 0.0000                                                           & 0.0350                                                            & \textbf{0.9567}& 1950.00             \\
\textbf{4}  & 1.0000                                                         & 1.0000                                                     & 1.0000      & 1.0000           & 1.0000                                                               & 1.0000                                                           & 1.0000                                                            & \textbf{1.0000}& 0.00                \\
\textbf{5}  & 0.0183                                                         & 0.0000                                                     & 0.0033      & 0.0300           & 0.0350                                                               & 0.0000                                                           & 0.0467                                                            & \textbf{0.2633}& 464.28              \\
\textbf{6}  & 1.0000                                                         & 1.0000                                                     & 1.0000      & 1.0000           & 1.0000                                                               & 1.0000                                                           & 1.0000                                                            & \textbf{1.0000}& 0.00                \\
\textbf{7}  & 1.0000                                                         & 1.0000                                                     & 1.0000      & 1.0000           & 1.0000                                                               & 1.0000                                                           & 1.0000                                                            & \textbf{1.0000}& 0.00                \\
\textbf{8}  & 0.8500                                                         & 0.6683                                                     & 0.8883      & 0.8500           & 0.8883                                                               & 0.7000                                                           & 0.9000                                                            & \textbf{0.9433}& 4.81                \\
\textbf{9}  & 0.0300                                                         & 0.0000                                                     & 0.0567      & 0.0633           & 0.0517                                                               & 0.0000                                                           & 0.0400                                                            & \textbf{0.8850}& 1297.37             \\
\textbf{10} & 0.6700                                                         & 0.5367                                                     & 0.8100      & 0.8033           & 0.8233                                                               & 0.0400                                                           & 0.8350                                                            & \textbf{0.8833}& 5.79                \\
\textbf{11} & 0.9400                                                         & 0.8417                                                     & 0.9433      & 0.9500           & 0.9483                                                               & 0.8600                                                           & 0.9433                                                            & \textbf{0.9533}& 0.35                \\
\textbf{12} & 0.3467                                                         & 0.0000                                                     & 0.3083      & 0.3267           & 0.4383                                                               & 0.0100                                                           & 0.4550                                                            & \textbf{0.6467}& 42.12               \\
\textbf{13} & 0.8467                                                         & 0.4167                                                     & 0.8483      & 0.8417           & 0.8583                                                               & 0.4133                                                           & 0.8583                                                            & \textbf{0.8733}& 1.75                \\
\textbf{14} & 0.9883                                                         & 0.6567                                                     & 0.9950      & 0.9967           & 0.9967                                                               & 0.7833                                                           & 0.9767                                                            & \textbf{0.9967}& 0.00                \\
\textbf{15} & 0.0117                                                         & 0.0000                                                     & 0.0150      & 0.0100           & 0.0133                                                               & 0.0000                                                           & 0.0317                                                            & \textbf{0.9850}& 3010.52             \\
\textbf{16} & 0.0167                                                         & 0.0000                                                     & 0.0117      & 0.0083           & 0.0083                                                               & 0.0000                                                           & 0.0250                                                            & \textbf{0.6450}& 2480.00             \\
\textbf{17} & 0.9017                                                         & 0.7683                                                     & 0.9083      & 0.9100           & 0.9067                                                               & 0.8283                                                           & 0.9050                                                            & \textbf{0.9233}& 2.03                \\
\textbf{18} & 0.3500                                                         & 0.2567                                                     & 0.4850      & 0.4817           & 0.4200                                                               & 0.3867                                                           & 0.3950                                                            & \textbf{0.7633}& 57.39               \\
\textbf{19} & 0.9317                                                         & 0.8000                                                     & 0.9783      & 0.9683           & 0.9683                                                               & 0.6550                                                           & 0.9750                                                            & \textbf{0.9817}& 0.68                \\
\textbf{20} & 0.7650                                                         & 0.5400                                                     & 0.7933      & 0.7900           & 0.7750                                                               & 0.5583                                                           & 0.7917                                                            & \textbf{0.7933}& 0.00                \\ \bottomrule
\end{tabular}
\end{adjustbox}
\end{table}

\begin{table}[H] \caption{Comparison of false alarm rates in TEP.\label{tab6}}
\centering \vspace{1em}
\begin{adjustbox}{width=\columnwidth}
\begin{tabular}{ccccccccc}
\toprule
\textbf{}    & \textbf{\begin{tabular}[c]{@{}c@{}}PCA\\ ($T^2$)\end{tabular}} & \textbf{\begin{tabular}[c]{@{}c@{}}PCA\\ (Q)\end{tabular}} & \textbf{AE} & \textbf{LSTM AE} & \textbf{\begin{tabular}[c]{@{}c@{}}Block PCA\\ ($T^2$)\end{tabular}} & \textbf{\begin{tabular}[c]{@{}c@{}}Block PCA\\ (Q)\end{tabular}} & \textbf{\begin{tabular}[c]{@{}c@{}}Block \\ LSTM AE\end{tabular}} & \textbf{MOLA} \\ \midrule
\textbf{FDR} & 0.0017                                                         & 0.0000                                                     & 0.0133      & 0.0067           & 0.0067                                                               & 0.0000                                                           & 0.0283                                                            & 0.0350        \\ \bottomrule
\end{tabular}
\end{adjustbox}
\end{table}

\subsection{Ablation studies evaluating the contribution of MOLA components to its process monitoring performance}

Our MOLA framework consists of several innovative components. To examine how these components contribute to the overall success of MOLA, we present results from ablation studies designed to individually evaluate the effectiveness of each component or improvement, offering quantitative understanding and deep insights into MOLA.

We first validate the effectiveness of MOLA in extracting non-redundant features. The Maximal Information Coefficient (MIC) \cite{kinney2014equitability}, ranging from 0 to 1, serves as a quantitative measure of the correlation between two variables. A MIC value of 0 between two variables indicates their mutual independence, whereas as MIC value approaches 1 it suggests the presence of a strong correlation between the two variables. Here, we use MIC as an indicator to assess the correlation among the extracted features. Specifically, we calculate the MIC values between any two features extracted by both the LSTM AE and the MOLA and summarize the results in Table  \ref{tab7}. Clearly, the MIC values for features extracted by MOLA are close to 0, whereas the MIC values for features extracted by LSTM AE are significantly larger. This indicates that MOLA demonstrates a significant advantage in extracting non-redundant features. As illustrated in Table \ref{tab8}, this leads to faster fault detection speed at the block level for MOLA compared to LSTM AE, when both methods use $T^2$ as the monitoring statistic. Meanwhile, MOLA without W-BF and CUSUM can detect process faults earlier than LSTM AE. This result indicates that introducing orthogonality constraints can facilitate faster detection of process faults. This is primarily because the redundancies among features extracted by traditional autoencoders may obscure the underlying data structure, making faults harder to detect. Instead, our proposed OLAE only extracts orthogonal features, making fault information more prominent and thus easier to detect.

\begin{table}[H] \caption{Orthogonality and mutual independence of extracted latent variables. \label{tab7}}
\centering \vspace{1em}
\begin{adjustbox}{width=\columnwidth}
\begin{tabular}{ccccccccccc}
\toprule
\multirow{2}{*}{\textbf{No.}} & \multicolumn{2}{c}{\textbf{Block 1}} & \multicolumn{2}{c}{\textbf{Block 2}} & \multicolumn{2}{c}{\textbf{Block 3}} & \multicolumn{2}{c}{\textbf{Block 4}} & \multirow{2}{*}{\textbf{\begin{tabular}[c]{@{}c@{}}Block \\ LSTM AE\end{tabular}}} & \multirow{2}{*}{\textbf{\begin{tabular}[c]{@{}c@{}}MOLA w/o \\ BF and CUSUM\end{tabular}}} \\
& \textbf{LSTM AE}   & \textbf{OLAE}   & \textbf{LSTM AE}   & \textbf{OLAE}   & \textbf{LSTM AE}   & \textbf{OLAE}   & \textbf{LSTM AE}   & \textbf{OLAE}   &                                                                                    &                                                                                           \\ \midrule
\textbf{1}                    & 21                 & \textbf{18}     & 4                  & \textbf{1}      & 4                  & \textbf{3}      & 3                  & \textbf{2}      & 4& \textbf{2}\\
\textbf{2}                    & \textbf{117}       & 118             & 21                 & \textbf{17}     & 22                 & \textbf{16}     & 18                 & \textbf{9}      & 18& \textbf{16}\\
\textbf{3}                    & -                  & -               & -                  & \textbf{139}    & 573                & \textbf{573}    & -                  & -               & 572& \textbf{160}\\
\textbf{4}                    & -                  & -               & 0                  & \textbf{0}      & 10                 & \textbf{10}     & -                  & \textbf{44}     & 0& \textbf{0}\\
\textbf{5}                    & -                  & -               & -                  & \textbf{1}      & 3                  & \textbf{2}      & -                  & \textbf{3}      & 3& \textbf{2}\\
\textbf{6}                    & 0                  & \textbf{0}      & 1                  & \textbf{0}      & 3                  & \textbf{3}      & 2                  & \textbf{2}      & 0& \textbf{0}\\
\textbf{7}                    & -                  & -               & 0                  & \textbf{0}      & 1                  & \textbf{1}      & 0                  & \textbf{0}      & 0& \textbf{0}\\
\textbf{8}                    & 56                 & \textbf{36}     & 182                & \textbf{28}     & 32                 & \textbf{32}     & 27                 & \textbf{24}     & 27& \textbf{27}\\
\textbf{9}                    & -                  & \textbf{23}     & 233                & \textbf{61}     & -                  & \textbf{61}     & -                  & \textbf{62}     & 233& \textbf{61}\\
\textbf{10}                   & -                  & -               & -                  & \textbf{380}    & 170                & \textbf{170}    & 74                 & \textbf{70}     & 74& \textbf{70}\\
\textbf{11}                   & -                  & -               & 36                 & \textbf{36}     & 47                 & \textbf{41}     & 63                 & \textbf{49}     & 36& \textbf{36}\\
\textbf{12}                   & -                  & \textbf{553}    & 85                 & \textbf{60}     & \textbf{48}        & 60              & 65                 & \textbf{65}     & \textbf{48}& 60\\
\textbf{13}                   & 249                & \textbf{245}    & 97                 & \textbf{74}     & 89                 & \textbf{84}     & 106                & \textbf{9}      & 91& \textbf{84}\\
\textbf{14}                   & -                  & \textbf{202}    & 2                  & \textbf{2}      & -                  & \textbf{56}     & -                  & \textbf{26}     & 2& \textbf{2}\\
\textbf{15}                   & -                  & -               & -                  & \textbf{82}     & 192                & \textbf{191}    & -                  & \textbf{209}    & \textbf{191}& 192\\
\textbf{16}                   & -                  & -               & -                  & -               & 54                 & \textbf{18}     & -                  & \textbf{19}     & -& \textbf{25}\\
\textbf{17}                   & -                  & -               & 56                 & \textbf{54}     & 118                & \textbf{117}    & 140                & \textbf{57}     & 57& \textbf{56}\\
\textbf{18}                   & -                  & -               & 345                & \textbf{345}    & 263                & \textbf{260}    & 346                & \textbf{346}    & 268& \textbf{260}\\
\textbf{19}                   & -                  & -               & 409                & \textbf{244}    & 25                 & \textbf{20}     & 12                 & \textbf{11}     & 12& \textbf{12}\\
\textbf{20}                   & 164                & \textbf{162}    & 212                & \textbf{177}    & 129                & \textbf{129}    & \textbf{123}       & 124             & \textbf{123}& 127\\ \bottomrule
\end{tabular}
\end{adjustbox}
\noindent{\footnotesize{* ``0'' indicates that the fault is detected at the time of its introduction, and ``--'' indicates that the fault is not effectively detected during the entire monitoring period.}}
\end{table}

\begin{table}[H] \caption{Comparison of fault detection speed using LSTM AE and MOLA. \label{tab8}}
\centering \vspace{1em}
\begin{adjustbox}{width=\columnwidth}
\begin{tabular}{ccccccccc}
\toprule
\textbf{Fault} & \multicolumn{2}{c}{\textbf{Block 1}} & \multicolumn{2}{c}{\textbf{Block 2}} & \multicolumn{2}{c}{\textbf{Block 3}} & \multicolumn{2}{c}{\textbf{Block 4}} \\ 
\textbf{No.} & \textbf{LSTM AE}   & \textbf{MOLA}   & \textbf{LSTM AE}   & \textbf{MOLA}   & \textbf{LSTM AE}   & \textbf{MOLA}   & \textbf{LSTM AE}   & \textbf{MOLA}   \\ \midrule
\textbf{1}                    & 21& \textbf{18}& 4& \textbf{1}& 4& \textbf{3}& 3& \textbf{2}\\
\textbf{2}                    & \textbf{117}& 118& 21& \textbf{17}& 22& \textbf{16}& 18& \textbf{9}\\
\textbf{3}                    & --                  & --               & --                  & \textbf{139}& 573& \textbf{573}& \textbf{--}         & --               \\
\textbf{4}                    & --                  & --               & 0& \textbf{0}& 10& \textbf{10}& \textbf{--}         & \textbf{44}\\
\textbf{5}                    & --                  & --               & --                  & \textbf{1}& 3& \textbf{2}& \textbf{--}         & \textbf{3}\\
\textbf{6}                    & 0& \textbf{0}& 1& \textbf{0}& 3& \textbf{3}& 2& \textbf{2}\\
\textbf{7}                    & --                  & --               & 0& \textbf{0}& 1& \textbf{1}& 0& \textbf{0}\\
\textbf{8}                    & 56& \textbf{36}& 182& \textbf{28}& 32& \textbf{32}& 27& \textbf{24}\\
\textbf{9}                    & --                  & \textbf{23}& 233& \textbf{61}& --                  & \textbf{61}& --& \textbf{62}\\
\textbf{10}                   & --                  & --               & --                  & \textbf{380}& 170& \textbf{170}& 74& \textbf{70}\\
\textbf{11}                   & --                  & --               & 36& \textbf{36}& 47& \textbf{41}& 63& \textbf{49}\\
\textbf{12}                   & --                  & \textbf{1553}& 85& \textbf{60}& \textbf{48}& 60& 65& \textbf{65}\\
\textbf{13}                   & 249& \textbf{245}& 97& \textbf{74}& 89& \textbf{84}& 106& \textbf{9}\\
\textbf{14}                   & --                  & \textbf{202}& 2& \textbf{2}& --                  & \textbf{56}& --& \textbf{26}\\
\textbf{15}                   & --                  & --               & --                  & \textbf{82}& 192& \textbf{191}& --& \textbf{209}\\
\textbf{16}                   & --                  & --               & --                  & --               & 54& \textbf{18}& --& \textbf{19}\\
\textbf{17}                   & --                  & --               & 56& \textbf{54}& 118& \textbf{117}& 140& \textbf{57}\\
\textbf{18}                   & --                  & --               & 345& \textbf{345}& 263& \textbf{260}& 346& \textbf{346}\\
\textbf{19}                   & --                  & --               & 409& \textbf{244}& 25& \textbf{20}& 12& \textbf{11}\\
\textbf{20}                   & 164& \textbf{162}& 212& \textbf{177}& 129& \textbf{129}& \textbf{123}& 124\\ \bottomrule
\end{tabular}
\end{adjustbox}
\noindent{\footnotesize{* ``0'' indicates that the fault is detected at the time of its introduction, and ``--'' indicates that the fault is not effectively detected during the entire monitoring period.}}
\end{table}

Next, we investigate the benefits of introducing the quantile-based CUSUM method and adaptive W-BF strategy on process monitoring performance. By designing ablation experiments, we quantify the enhancements and present the results in Table \ref{table_TEPvariables0}. Specifically, ``Full MOLA'' represents the complete MOLA framework presented earlier, ``MOLA no CUSUM'' means the MOLA framework without implementing the CUSUM procedure, and ``MOLA no BF'' means the MOLA framework without implementing adaptive W-BF strategy. Table \ref{table_TEPvariables0} shows that introducing adaptive W-BF significantly improves the fault detection speed for faults 13 and 15 and enhances fault detection rate for all faults other than 1, 4, 6, 7, 10, and 14. For instance, for fault 13, the weights assigned to blocks 1 through 4 by adaptive W-BF strategy at the 609th time step (i.e., 9 samples after fault is introduced) are 0.2396, 0.1372, 0.1542, and 0.4689, respectively. Block 4, which has the highest weight to PFI among all blocks, also turns out to be the first block where an alarm is raised. Thus, the adaptive W-BF method automatically prompts the FPI to give higher attention to blocks that detect anomalies earlier, which enhances both fault detection rate and speed.

 We remark that the W-BF method is particularly attractive in monitoring modern, large-scale industrial processes, whose scale and complexity require process monitoring to be performed in a distributed manner. Since process faults typically occurs locally, by decomposing the overall process into multiple blocks, our W-BF method ensures that, as a fault occurs, process monitoring performance is sensitive to the faulty blocks and will not be negatively impacted by the surveillance of non-faulty blocks.

\begin{table}[H]\caption{Contribution of quantile-based CUSUM and adaptive W-BF to improvements of process monitoring performance.} \label{table_TEPvariables0}
\centering \vspace{1em}
\begin{tabular}{ccccccc}
\toprule
\multirow{2}{*}{\textbf{Fault No.}} & \multicolumn{2}{c}{\textbf{Full MOLA}}      & \multicolumn{2}{c}{\textbf{MOLA no BF}} & \multicolumn{2}{c}{\textbf{MOLA no CUSUM}} \\
& \textbf{FDD}& \textbf{FDR}    & \textbf{FDD}& \textbf{FDR} & \textbf{FDD}& \textbf{FDR} \\ \midrule
\textbf{1}                    & \textbf{1}& \textbf{0.9983} & 1& 0.9983       & 1& 0.9983       \\
\textbf{2}                    & \textbf{16}& \textbf{0.9767} & 16& 0.9750       & 16& 0.9750       \\
\textbf{3}                    & \textbf{30}& \textbf{0.9567} & 31& 0.9550       & 139& 0.1117       \\
\textbf{4}                    & \textbf{0}& \textbf{1.0000} & 0& 1.0000       & 0& 1.0000       \\
\textbf{5}                    & \textbf{2}& \textbf{0.2633} & 2& 0.2283       & 2& 0.0717       \\
\textbf{6}                    & \textbf{0}& \textbf{1.0000} & 0& 1.0000       & 0& 1.0000       \\
\textbf{7}                    & \textbf{0}& \textbf{1.0000} & 0& 1.0000       & 0& 1.0000       \\
\textbf{8}                    & \textbf{25}& \textbf{0.9433} & 25& 0.9350       & 26& 0.9167       \\
\textbf{9}                    & 25& \textbf{0.8850} & 25& 0.8800       & \textbf{23}& 0.1550       \\
\textbf{10}                   & \textbf{70}& \textbf{0.8833} & 70& 0.8833       & 70& 0.8567       \\
\textbf{11}                   & \textbf{36}& \textbf{0.9533} & 36& 0.9500       & 36& 0.9500       \\
\textbf{12}                   & \textbf{60}& \textbf{0.6467} & 60& 0.6217       & 60& 0.5517       \\
\textbf{13}                   & \textbf{09}& \textbf{0.8733} & 84& 0.8667       & 84& 0.8717       \\
\textbf{14}                   & \textbf{2}& \textbf{0.9967} & 2& 0.9967       & 2& 0.9967       \\
\textbf{15}                   & \textbf{6}& \textbf{0.9850} & 12& 0.9550       & 192& 0.0667       \\
\textbf{16}                   & \textbf{21}& \textbf{0.6450} & 25& 0.6233       & 25& 0.0450       \\
\textbf{17}                   & \textbf{24}& \textbf{0.9233} & 24& 0.9183       & 54& 0.9117       \\
\textbf{18}                   & \textbf{147}& \textbf{0.7633} & 148& 0.7567       & 260& 0.4567       \\
\textbf{19}                   & \textbf{11}& \textbf{0.9817} & 12& 0.9800       & 11& 0.9817       \\
 \textbf{20}& \textbf{124}& \textbf{124}& 124& 0.7917& 124&0.7933\\
\textbf{Average}& \textbf{30.45}& \textbf{0.8734}& 30.85& 0.8658& 56.25& 0.6855\\ \bottomrule
\end{tabular}
\end{table}

\begin{figure}[H]
\centering
\includegraphics[width=\textwidth]{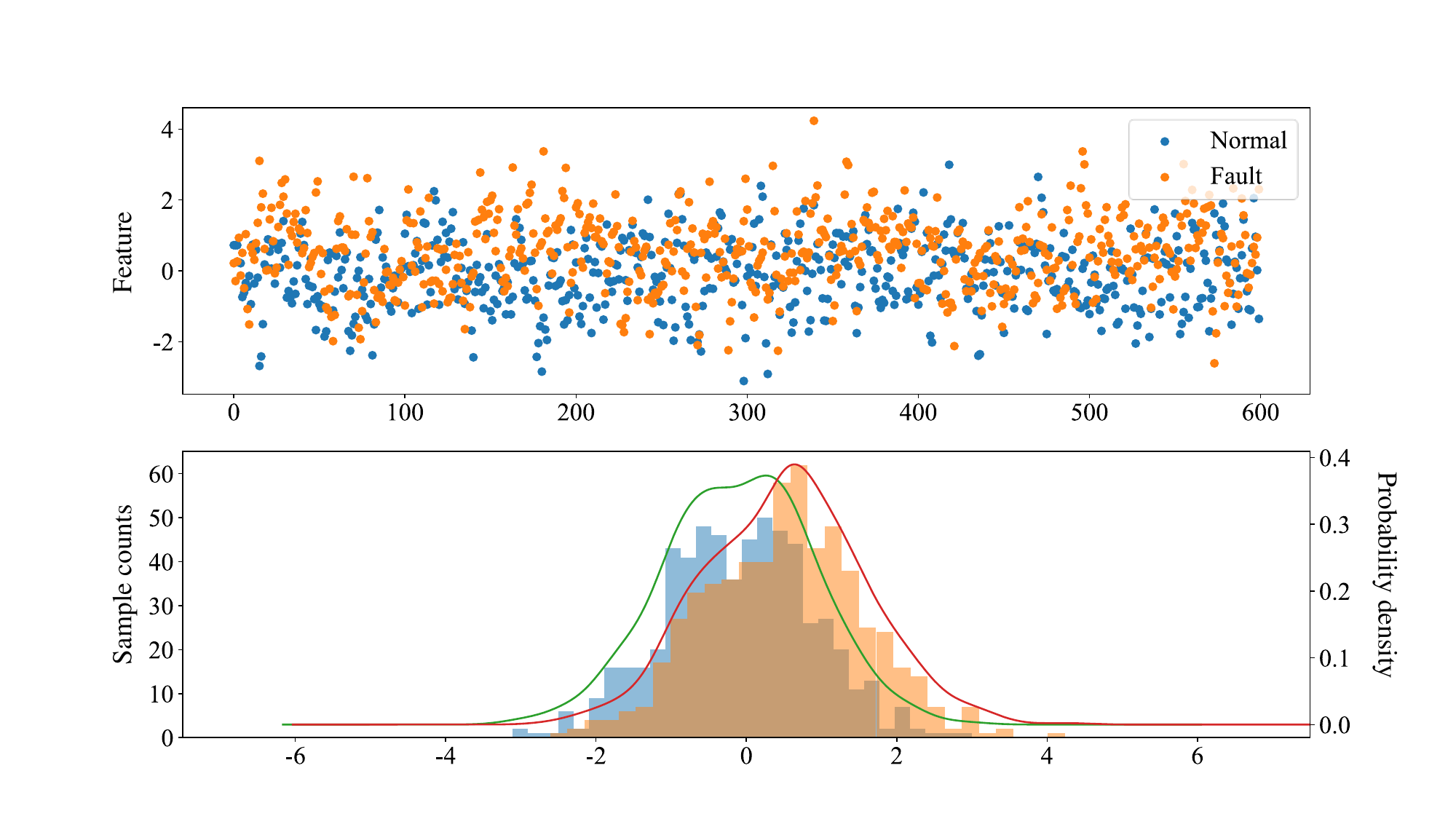}
\vspace{-3em}
\caption{Feature values and its distribution.\label{fig8}}
\end{figure}

Meanwhile, a comparison between ``full MOLA'' and ``MOLA no CUSUM'' shows that incorporating quantile-based CUSUM procedure greatly enhances the fault detection rate and speed for hard-to-detect faults of 3, 9, and 15. Again, take fault 3 as an example. As shown in Figure \ref{fig8}, under normal and faulty conditions, the numerical range of one of the specific features extracted from block 2 remains largely the same, making traditional distance-based monitoring statistics such as $T^2$ struggle to detect the fault. However, when plotting the distribution of the features under normal and faulty conditions, considerable changes are noticed. Therefore, the quantile-based multivariate CUSUM method, which directly targets the detection of distribution changes, significantly improves the accuracy and speed of fault detection under these scenarios. Furthermore, as shown in Figure \ref{table_TEPvariables0}, the quantile-based CUSUM statistic continuously accumulates as fault 3 occurs and will not go below the control limit once an alarm is raised. This causes less confusion among plant operators in interpreting alarms and thus offers an additional advantage over other monitoring methods.

\begin{figure}[H]
\centering
\includegraphics[width=\textwidth]{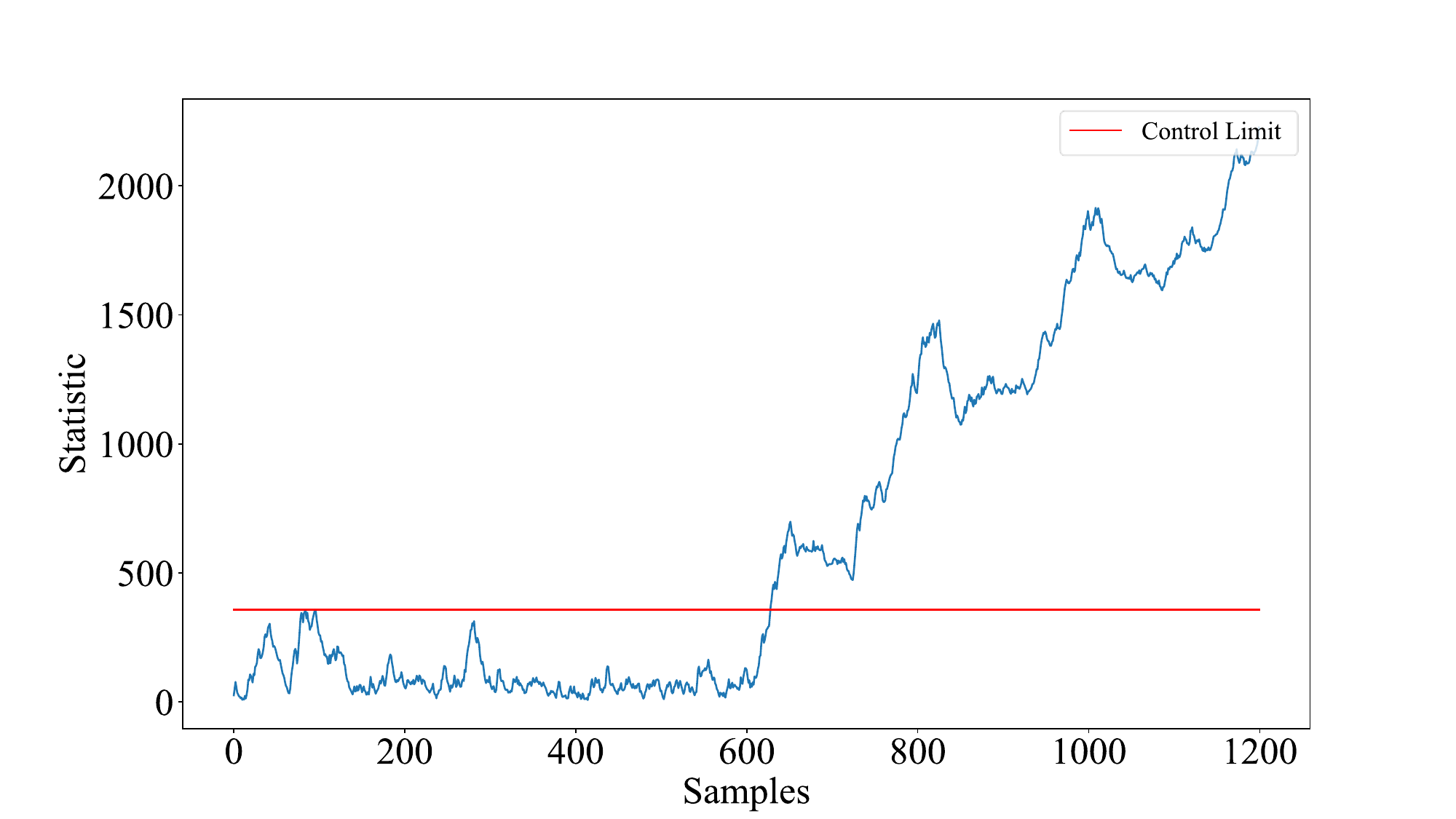}
\vspace{-2em}
\caption{Variations of the quantile-based non-parametric CUSUM statistic for block 2.\label{fig9}}
\end{figure}  

\section{Conclusions}

In this work, we develop a novel process monitoring framework named MOLA for large-scale industrial processes. By adopting a multi-block monitoring strategy, MOLA successfully addresses the challenges posed by the scale and complexity of modern industrial processes. MOLA can extract dynamic orthogonal latent features, making sure that the most essential, non-redundant features are identified and extracted. In addition, MOLA incorporates a quantile-based multivariate CUSUM method which enhances the ability to detect faults characterized by subtle changes in feature distributions. Furthermore, the adaptive weight-based Bayesian fusion strategy enhances fault detection rate and speed. We remark that these methods lead to synergistic improvement in process monitoring performance when they are integrated with the MOLA framework. Case study results on the TEP problem indicate that MOLA not only significantly improves fault detection rates and speeds but also successfully detects faults that are considered difficult to be detected in prior research, thereby opening up many exciting opportunities for fast, accurate, and reliable industrial process monitoring applications.

Despite these achievements, we remark that there is still room for improvement for our MOLA framework. For instance, the current block assignment technique is solely based on our expert knowledge without fully considering the correlations among blocks, which may have an impact on both local and global monitoring performance. Therefore, future research will focus on exploring more systematic methods for dividing process sub-blocks, taking into account the correlations between sub-blocks to further optimize the process monitoring model. For example, we plan to perform correlation analysis on process variables prior to assigning them to different blocks. This way, the dynamic features extracted not only capture the characteristics of process data within individual blocks, but also incorporate the interconnections across different blocks. This will further enhance its monitoring performance and robustness in complex industrial processes.


\bibliography{ref}

\begin{thebibliography}{10}
\expandafter\ifx\csname url\endcsname\relax
  \def\url#1{\texttt{#1}}\fi
\expandafter\ifx\csname urlprefix\endcsname\relax\def\urlprefix{URL }\fi
\expandafter\ifx\csname href\endcsname\relax
  \def\href#1#2{#2} \def\path#1{#1}\fi

\bibitem{amin2018process}
M.~T. Amin, S.~Imtiaz, F.~Khan, Process system fault detection and diagnosis using a hybrid technique, Chemical Engineering Science 189 (2018) 191--211.

\bibitem{nawaz2022review}
M.~Nawaz, A.~S. Maulud, H.~Zabiri, H.~Suleman, Review of multiscale methods for process monitoring, with an emphasis on applications in chemical process systems, IEEE Access 10 (2022) 49708--49724.

\bibitem{qin2012survey}
S.~J. Qin, Survey on data-driven industrial process monitoring and diagnosis, Annual reviews in control 36~(2) (2012) 220--234.

\bibitem{li2022nonlinear}
S.~Li, J.~Luo, Y.~Hu, Nonlinear process modeling via unidimensional convolutional neural networks with self-attention on global and local inter-variable structures and its application to process monitoring, ISA transactions 121 (2022) 105--118.

\bibitem{dong2018novel}
Y.~Dong, S.~J. Qin, A novel dynamic pca algorithm for dynamic data modeling and process monitoring, Journal of Process Control 67 (2018) 1--11.

\bibitem{bounoua2021fault}
W.~Bounoua, A.~Bakdi, Fault detection and diagnosis of nonlinear dynamical processes through correlation dimension and fractal analysis based dynamic kernel pca, Chemical Engineering Science 229 (2021) 116099.

\bibitem{pilario2019review}
K.~E. Pilario, M.~Shafiee, Y.~Cao, L.~Lao, S.-H. Yang, A review of kernel methods for feature extraction in nonlinear process monitoring, Processes 8~(1) (2019) 24.

\bibitem{tan2020monitoring}
R.~Tan, J.~R. Ottewill, N.~F. Thornhill, Monitoring statistics and tuning of kernel principal component analysis with radial basis function kernels, IEEE Access 8 (2020) 198328--198342.

\bibitem{abiodun2018state}
O.~I. Abiodun, A.~Jantan, A.~E. Omolara, K.~V. Dada, N.~A. Mohamed, H.~Arshad, State-of-the-art in artificial neural network applications: A survey, Heliyon 4~(11) (2018).

\bibitem{wu2018deep}
H.~Wu, J.~Zhao, Deep convolutional neural network model based chemical process fault diagnosis, Computers \& chemical engineering 115 (2018) 185--197.

\bibitem{arunthavanathan2021deep}
R.~Arunthavanathan, F.~Khan, S.~Ahmed, S.~Imtiaz, A deep learning model for process fault prognosis, Process Safety and Environmental Protection 154 (2021) 467--479.

\bibitem{heo2018fault}
S.~Heo, J.~H. Lee, Fault detection and classification using artificial neural networks, IFAC-PapersOnLine 51~(18) (2018) 470--475.

\bibitem{yang2022autoencoder}
Z.~Yang, B.~Xu, W.~Luo, F.~Chen, Autoencoder-based representation learning and its application in intelligent fault diagnosis: A review, Measurement 189 (2022) 110460.

\bibitem{ji2022review}
C.~Ji, W.~Sun, A review on data-driven process monitoring methods: Characterization and mining of industrial data, Processes 10~(2) (2022) 335.

\bibitem{fan2017autoencoder}
J.~Fan, W.~Wang, H.~Zhang, Autoencoder based high-dimensional data fault detection system, in: 2017 ieee 15th international conference on industrial informatics (indin), IEEE, 2017, pp. 1001--1006.

\bibitem{qian2022review}
J.~Qian, Z.~Song, Y.~Yao, Z.~Zhu, X.~Zhang, A review on autoencoder based representation learning for fault detection and diagnosis in industrial processes, Chemometrics and Intelligent Laboratory Systems 231 (2022) 104711.

\bibitem{wang2019clustering}
W.~Wang, D.~Yang, F.~Chen, Y.~Pang, S.~Huang, Y.~Ge, Clustering with orthogonal autoencoder, IEEE Access 7 (2019) 62421--62432.

\bibitem{cacciarelli2022novel}
D.~Cacciarelli, M.~Kulahci, A novel fault detection and diagnosis approach based on orthogonal autoencoders, Computers \& Chemical Engineering 163 (2022) 107853.

\bibitem{md2020review}
N.~Md~Nor, C.~R. Che~Hassan, M.~A. Hussain, A review of data-driven fault detection and diagnosis methods: Applications in chemical process systems, Reviews in Chemical Engineering 36~(4) (2020) 513--553.

\bibitem{ji2024orthogonal}
C.~Ji, F.~Ma, J.~Wang, W.~Sun, Orthogonal projection based statistical feature extraction for continuous process monitoring, Computers \& Chemical Engineering 183 (2024) 108600.

\bibitem{nawaz2021improved}
M.~Nawaz, A.~S. Maulud, H.~Zabiri, S.~A.~A. Taqvi, A.~Idris, Improved process monitoring using the cusum and ewma-based multiscale pca fault detection framework, Chinese Journal of Chemical Engineering 29 (2021) 253--265.

\bibitem{ge2013distributed}
Z.~Ge, Z.~Song, Distributed pca model for plant-wide process monitoring, Industrial \& engineering chemistry research 52~(5) (2013) 1947--1957.

\bibitem{huang2019fault}
J.~Huang, O.~K. Ersoy, X.~Yan, Fault detection in dynamic plant-wide process by multi-block slow feature analysis and support vector data description, ISA transactions 85 (2019) 119--128.

\bibitem{zhai2020multi}
C.~Zhai, X.~Sheng, W.~Xiong, Multi-block fault detection for plant-wide dynamic processes based on fault sensitive slow features and support vector data description, IEEE Access 8 (2020) 120737--120745.

\bibitem{li2020plant}
Y.~Li, X.~Peng, Y.~Tian, Plant-wide process monitoring strategy based on complex network and bayesian inference-based multi-block principal component analysis, IEEE Access 8 (2020) 199213--199226.

\bibitem{ye2022generic}
H.~Ye, K.~Liu, A generic online nonparametric monitoring and sampling strategy for high-dimensional heterogeneous processes, IEEE Transactions on Automation Science and Engineering 19~(3) (2022) 1503--1516.

\bibitem{jiang2023online}
Z.~Jiang, Online monitoring and robust, reliable fault detection of chemical process systems, in: Computer Aided Chemical Engineering, Vol.~52, Elsevier, 2023, pp. 1623--1628.

\bibitem{zhang2019bidirectional}
S.~Zhang, K.~Bi, T.~Qiu, Bidirectional recurrent neural network-based chemical process fault diagnosis, Industrial \& Engineering Chemistry Research 59~(2) (2019) 824--834.

\bibitem{deng2023lstmed}
W.~Deng, Y.~Li, K.~Huang, D.~Wu, C.~Yang, W.~Gui, Lstmed: An uneven dynamic process monitoring method based on lstm and autoencoder neural network, Neural Networks 158 (2023) 30--41.

\bibitem{ren2020batch}
J.~Ren, D.~Ni, A batch-wise lstm-encoder decoder network for batch process monitoring, Chemical Engineering Research and Design 164 (2020) 102--112.

\bibitem{ma2022data}
F.~Ma, J.~Wang, W.~Sun, A data-driven semi-supervised soft-sensor method: Application on an industrial cracking furnace, Frontiers in Chemical Engineering 4 (2022) 899941.

\bibitem{mao2018feature}
T.~Mao, Y.~Zhang, Y.~Ruan, H.~Gao, H.~Zhou, D.~Li, Feature learning and process monitoring of injection molding using convolution-deconvolution auto encoders, Computers \& Chemical Engineering 118 (2018) 77--90.

\bibitem{qiu2003nonparametric}
P.~Qiu, D.~Hawkins, A nonparametric multivariate cumulative sum procedure for detecting shifts in all directions, Journal of the Royal Statistical Society Series D: The Statistician 52~(2) (2003) 151--164.

\bibitem{mei}
Y.~Mei, Quickest detection in censoring sensor networks, in: 2011 IEEE International Symposium on Information Theory Proceedings, 2011, pp. 2148--2152.

\bibitem{rong2021multi}
M.~Rong, H.~Shi, B.~Song, Y.~Tao, Multi-block dynamic weighted principal component regression strategy for dynamic plant-wide process monitoring, Measurement 183 (2021) 109705.

\bibitem{ge2015plant}
Z.~Ge, J.~Chen, Plant-wide industrial process monitoring: A distributed modeling framework, IEEE Transactions on Industrial Informatics 12~(1) (2015) 310--321.

\bibitem{downs1993plant}
J.~J. Downs, E.~F. Vogel, A plant-wide industrial process control problem, Computers \& chemical engineering 17~(3) (1993) 245--255.

\bibitem{jiang2019review}
Q.~Jiang, X.~Yan, B.~Huang, Review and perspectives of data-driven distributed monitoring for industrial plant-wide processes, Industrial \& Engineering Chemistry Research 58~(29) (2019) 12899--12912.

\bibitem{zhu2017distributed}
J.~Zhu, Z.~Ge, Z.~Song, Distributed parallel pca for modeling and monitoring of large-scale plant-wide processes with big data, IEEE Transactions on Industrial Informatics 13~(4) (2017) 1877--1885.

\bibitem{bathelt2015revision}
A.~Bathelt, N.~L. Ricker, M.~Jelali, Revision of the tennessee eastman process model, IFAC-PapersOnLine 48~(8) (2015) 309--314.

\bibitem{alakent2024early}
B.~Alakent, Early fault detection via combining multilinear pca with retrospective monitoring using weighted features, Brazilian Journal of Chemical Engineering (2024) 1--23.

\bibitem{kinney2014equitability}
J.~B. Kinney, G.~S. Atwal, Equitability, mutual information, and the maximal information coefficient, Proceedings of the National Academy of Sciences 111~(9) (2014) 3354--3359.

\end{thebibliography}
\bibliographystyle{elsarticle-num}

\end{document}